\documentclass[runningheads]{llncs}

\usepackage{eccv}

\usepackage{marvosym}
\usepackage{eccvabbrv}
\usepackage{color}
\usepackage{graphicx}
\usepackage{booktabs}
\usepackage{wrapfig}
\usepackage[accsupp]{axessibility}  % Improves PDF readability for those with disabilities.

\usepackage{hyperref}
\usepackage{url}
\usepackage{multicol}
\usepackage{aliascnt}
\usepackage{adjustbox}
\usepackage{tcolorbox}
\usepackage{booktabs}
\usepackage{multirow} 
\usepackage{enumitem}
\usepackage{array}
\usepackage{footnote}
\usepackage{tablefootnote}
\usepackage{pifont}
\usepackage{makecell}
\usepackage{CJKutf8}
\definecolor{uclablue}{rgb}{0.15, 0.45, 0.68}
\hypersetup{
    breaklinks,
    citecolor=uclablue,
    colorlinks=true,
}
\usepackage{xcolor}

\usepackage{hyperref}

\usepackage{orcidlink}

\begin{document}

%\title{Deeply Bound Audiovisual Captioner for Multimodal Understanding and Generation}
%\title{ Improving Audiovisual Video Captions with Cross-Modal Binding}
 \title{Temporal and Cross-Modal Alignment for Enhanced Audiovisual Video Captioning} 

\titlerunning{Temporal and Cross-Modal Alignment for AVC}


% \author{Chen Zhao\inst{1,2}\orcidlink{0009-0000-3762-3728} \and
% Jiajun Ma\inst{1} \and
% Qilong Huang\inst{2} \and
% Tiehan Fan \inst{1} \and
% Hongyu Li \inst{2} \and
% Zhuoliang Kang \inst{2} \and
% Xiaoming Wei \inst{2} \and
% Jian Yang \inst{1} \and
% Ying Tai \inst{1} 
% }

\author{Chen Zhao\inst{1,2}\thanks{Equal contribution. This work was done while Chen Zhao is an intern at Meituan.%This work was conducted during Chen Zhao's internship at Meituan. %under the guidance of Zhuoliang Kang and Qilong Huang.
}\orcidlink{0009-0000-3762-3728} \and
Jiajun Ma\inst{1}$^{\star}$ \and
Qilong Huang\inst{2} \and
Tiehan Fan \inst{1} \and
Hongyu Li \inst{2} \and
Zhuoliang Kang \inst{2} \and
Xiaoming Wei \inst{2} \and
Jian Yang \inst{1} \and
Ying Tai \inst{1}$^{\dagger}$ 
% 如果你想用信封，这里可以换成 \inst{1}\textsuperscript{\Letter} (需引入 marvosym 宏包)
}

\authorrunning{C.~Zhao et al.}

\institute{Nanjing University, State Key Laboratory for Novel Software Technology, China 
%School of Intelligence Science and Technology, China 
\and
Meituan, China
\\
% \url{https://huggingface.co/collections/26TCA/eccv-2026-tca-captioner}
\href{https://huggingface.co/collections/26TCA/eccv-2026-tca-captioner}{https://huggingface.co/TCA-Captioner-ECCV-2026}
}

\maketitle
{
\renewcommand{\thefootnote}{\relax}
\footnotetext{$^{\dagger}$ Corresponding author.}
}

\begin{abstract}
While Multimodal Large Language Models (MLLMs) have advanced video understanding, achieving precise temporal and cross-modal alignment in audiovisual video captioning remains a formidable challenge. Most existing approaches suffer from modality detachment and temporal incoherence, failing to accurately bind auditory events to visual entities or capture complex causal dynamics. To address these deficiencies, we propose TCA-Captioner, a framework specifically engineered to enhance Temporal and Cross-Modal Alignment for  audiovisual video captioning. We first introduce the Observer-Checker-Corrector (OCC) framework, an iterative refinement strategy that generates high-fidelity, meticulously grounded training data. Leveraging a curated high-density human interaction dataset, TCA-Captioner is optimized to model sophisticated audiovisual interactions. Furthermore, we present TCA-Bench, a diagnostic benchmark utilizing a Decoupled Evaluation Protocol to isolate and quantify model proficiency in audiovisual binding and temporal relational reasoning. Extensive experiments demonstrate that TCA-Captioner sets a new standard for temporally-coherent and synchronized audiovisual narratives.

  \keywords{Video Captioning \and MLLM \and LLM}
\end{abstract}

\section{Introduction}
\label{sec:intro}

 %(e.g., misattributing speech or sounds to the wrong visual source)(e.g., chronological inversions or unsynchronized event sequences)
\begin{figure}[h]
	\centering
	\includegraphics[width=1\linewidth]{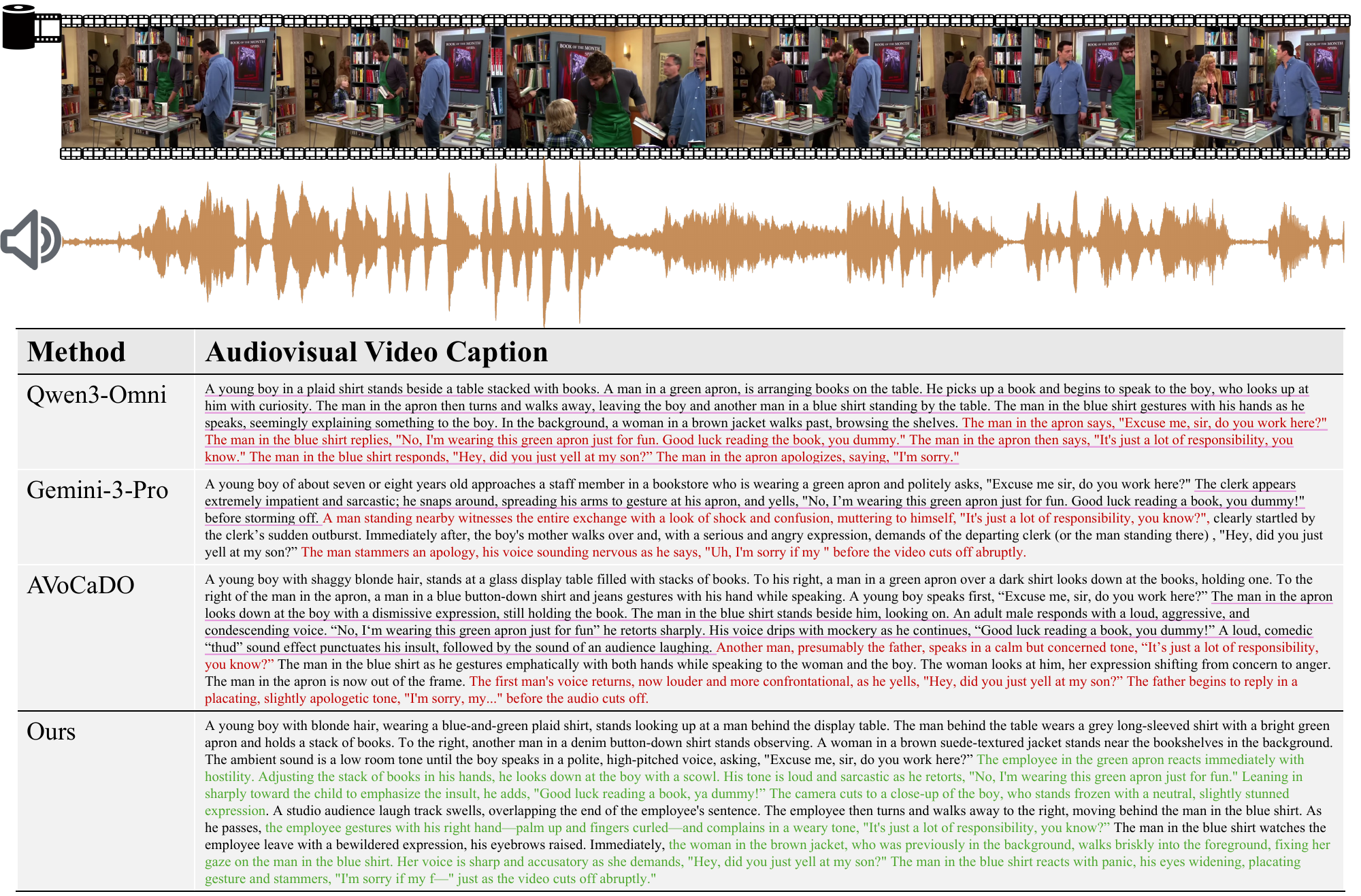}
 \vspace{-0.5cm}
	\caption{The challenges for audiovisual video captioning. Existing state-of-the-art models, including Qwen3-Omni, Gemini-3-Pro, and AVoCaDO, still struggle with two fundamental challenges: \textbf{Audio-Visual Cross-Modal Binding} and \textbf{Audio-Visual Temporal Coherence}. In the provided examples, text highlighted in {\color{red}red} indicates errors in Cross-Modal Binding. Text marked with {\color{Orchid} \underline{purple underlines}} identifies failures in Temporal Coherence. In contrast, our model, shown in the bottom row, accurately captures these complex interactions, with correctly grounded and synchronized descriptions highlighted in {\color{green}green}. The original video file will be provided in the supplementary material to facilitate a more detailed comparison and observation.} 
	
 \vspace{-0.5cm}
	\label{fig:motivation}
\end{figure}

Amidst the rapid ascendance of Multimodal Large Language Models (MLLMs) \cite{wang2024qwen2,zhu2023minigpt,zhang2023video,zhang2023speechgpt,liu2026driveva,lu2024robust,lu2025does,zhou2025bidedpo,zhou2026metapoint}, video captioning plays a pivotal role in advancing the frontiers of video understanding and generation\cite{du2025vc4vg,chen2024sharegpt4video,fan2025instancecap,lian2025describe,zhao2026learning,zhao2025ultrahr,lu2024mace,lu2023tf,zhou20243dis,zhou2025dreamrenderer,zhang2026probing,zhangrethinking,huang2026expo}. High-quality video captions serve not only as a foundational cornerstone for cross-modal alignment during pre-training but also as a critical mechanism for injecting rich semantic knowledge into downstream tasks \cite{qasim2025dense, guo2026alive, team2025longcat,chen2026calibrated,chen2026l2punlockinglatentpotential,chen2025dip,11513016,Zhaowfdiff}. Recent empirical studies suggest that training with detailed and factually grounded captions yields consistent performance gains across a broad spectrum of applications, including video question answering\cite{kim2025vru}, action recognition\cite{wang2025haic}, and generative video synthesis\cite{guo2026alive,nan2025openvid}. Therefore, enhancing the capabilities of video captioning models has emerged as a fundamental pathway toward building more robust  video understanding and generation systems.
%and versatile

Despite the remarkable progress of MLLMs, early research in video captioning has remained predominantly vision-centric\cite{yuan2025tarsier2,chen2024sharegpt4video,fan2025instancecap}, often overlooking the profound semantic cues embedded within auditory signals. In practice, auditory elements such as dialogues, voiceovers, and ambient soundscapes are indispensable for achieving a holistic and embodied understanding of video content. The emergence of omni-modal MLLMs, such as Qwen3-Omni\cite{xu2025qwen3} and Gemini 3 Pro \cite{gemini}, has signaled a paradigm shift toward joint audio-visual reasoning. Concurrently, specialized audiovisual captioners, including AVoCaDO\cite{chen2025avocado}, video-SALMONN 2\cite{tang2025video}, and UGC-VideoCaptioner\cite{wu2025ugc}, have begun to explore the complementary nature of auditory and visual streams to produce more comprehensive narratives.

Nevertheless, two critical challenges in joint audio-visual video captioning remain insufficiently addressed. The first is \textbf{Audio-Visual Cross-Modal Binding}, which refers to the precise association between specific auditory events and their corresponding visual entities. For instance, models frequently struggle to correctly attribute a segment of speech to its specific speaker in crowded scenes or link a transient impact sound to its physical origin, leading to attribute misattribution. The second challenge is \textbf{Audio-Visual Temporal Coherence}, which concerns the chronological ordering and causal dependencies across audio-visual modalities. Existing models often fail to capture the fine-grained temporal dynamics between audio signals and visual actions, such as the "sound-before-sight" anticipation or "action-followed-by-sound" feedback, thereby struggling to model the evolving logic of multimodal event streams. As illustrated in Fig. \ref{fig:motivation}, Qwen3-Omni\cite{xu2025qwen3} exhibits a pronounced modality isolation in its descriptions; it tends to narrate visual content and auditory elements in a decoupled, sequential manner (i.e., describing all visual scenes before addressing any audio), which entirely neglects the temporal coherence between the two streams. Furthermore, it suffers from significant errors in cross-modal binding. Similarly, both Gemini-3-Pro\cite{gemini} and AVoCaDO \cite{chen2025avocado} demonstrate localized failures in binding and temporal synchronization within their generated captions.
These deficiencies represent a significant barrier to achieving a truly unified perception of synchronized audio-visual content.

To bridge these gaps, we present TCA-Captioner along with a diagnostic benchmark, TCA-Bench, specifically engineered to enhance \textbf{T}emporal and \textbf{C}ross-modal \textbf{A}lignment in audiovisual video captioning.
To facilitate high-fidelity model training, we first design a rigorous data synthesis pipeline termed Observer-Checker-Corrector (OCC), powered by the state-of-the-art Gemini 3 Pro. Unlike conventional single-pass annotation, the OCC pipeline adopts an iterative refinement strategy. In this framework, the Observer first generates initial descriptions, which are then rigorously vetted by the Checker to identify unimodal factual hallucinations and cross-modal alignment inconsistencies. By ensuring the narrative is faithful to both individual streams and their joint dynamics, this verification step allows the Corrector to precisely rectify misalignments, thereby producing meticulously grounded captions.
Furthermore, to specifically address the intricacies of temporal and cross-modal dependencies, we curate a  high-density human interaction Dataset. This dataset is characterized by dense human speech interleaved with complex physical actions, providing a rich corpus of synchronized audiovisual cues that are often absent in existing datasets. Leveraging this high-fidelity data, we optimize TCA-Captioner through a robust post-training regime centered on Supervised Fine-Tuning (SFT). Extensive experiments demonstrate that our TCA-Captioner achieves superior performance in modeling sophisticated cross-modal interactions, effectively setting a new standard for temporally-coherent audiovisual narratives.

To facilitate a more granular assessment of audiovisual video captioning, we introduce TCA-Bench, a diagnostic benchmark that advocates a paradigm shift from static detail recall to the evaluation of dynamic audio-visual entanglement. Unlike existing benchmarks that rely on holistic scoring or "bag-of-events" matching, which frequently neglect the structural coherence of narratives, TCA-Bench specifically addresses the prevalent failure mode of modality detachment. We provide structured annotations, including Audio-Visual Binding Lists and Temporal Relational Lists, to formalize the fine-grained dependencies between acoustic events and their visual origins. Furthermore, we propose a Decoupled Evaluation Protocol that disentangles foundational uni-modal perception from higher-order relational reasoning. This protocol employs an LLM-based judge to perform targeted, hallucination-resistant verification of cross-modal source localization and chronological ordering. By isolating these synchronization capabilities, TCA-Bench offers a highly interpretable and granular assessment, uncovering the primary bottlenecks of current models in achieving unified and temporally-coherent multimodal understanding. Our contributions can be summarized as follows:

\begin{itemize}
\item We introduce the Observer-Checker-Corrector (OCC) framework, which employs an iterative refinement strategy to disentangle multifaceted audiovisual caption generation, guaranteeing the production of meticulously grounded and high-fidelity video descriptions.
%\item We curate a high-density human interaction dataset with synchronized audiovisual annotations, upon which we develop TCA-Captioner, a model optimized for fine-grained temporal and cross-modal alignment.
\item We develop TCA-Captioner, specifically optimized for fine-grained cross-modal binding and temporal coherence, supported by a newly curated dataset of high-density human interactions.
\item We propose TCA-Bench, a diagnostic benchmark utilizing a Decoupled Evaluation Protocol to  quantify the model's proficiency in audio-visual binding and temporal relational reasoning.
\end{itemize}
%By curating high-density interaction videos from YouTube, we provide structured annotations,

 %This process enables the model to transition from coarse scene recognition to fine-grained event synchronization.
%Unlike conventional single-pass annotation, the OCC pipeline employs an iterative refinement strategy where the Observer generates initial descriptions, the Checker identifies cross-modal inconsistencies, and the Corrector rectifies misalignments, thereby ensuring the production of meticulously grounded captions.

\section{Related Works}

\subsection{Multimodal Large Language Models}
The landscape of Multimodal Large Language Models (MLLMs) has undergone a rapid paradigm shift, expanding from basic image-text alignment to sophisticated, all-encompassing sensory perception \cite{wang2025internvl3}. Early endeavors, such as LLaVA\cite{liu2024improved} and BLIP-2 \cite{li2023blip}, focused on bridging the modality gap by aligning frozen visual encoders with LLMs via specialized adapters, a foundation later scaled by InternVL\cite{chen2024internvl} and VILA\cite{lin2024vila}. This evolved into temporal modeling with Video-LLaVA\cite{lin2024video}, ShareGPT4Video\cite{chen2024sharegpt4video}, and LLaVA-Video\cite{zhang2024llava}, which utilized high-quality synthetic data to capture dynamic transitions in silent videos. Concurrently, dedicated models like SALMONN\cite{tang2023salmonn} and Qwen2-Audio\cite{chu2024qwen2} emerged to tackle non-visual acoustic reasoning and spatial audio understanding\cite{zhang2023speechgpt}.
Building upon these modular successes, the field is converging toward Omni-modal MLLMs \cite{xu2025qwen3,gemini}, aiming for a holistic interpretation of real-world scenarios by processing interleaved text, vision, and audio tokens within a unified architecture. Industry leaders like GPT-4o\cite{hurst2024gpt} and the Gemini series (including Gemini 3 Pro)\cite{gemini} have demonstrated the efficacy of native multimodal integration, exhibiting superior reasoning in long-context and real-time environments. In the open-source community, the Qwen-Omni series\cite{xu2025qwen25omnitechnicalreport,xu2025qwen3} has mirrored this trend: while Qwen2.5-Omni\cite{xu2025qwen25omnitechnicalreport} achieved seamless audio-visual streaming, the state-of-the-art Qwen3-Omni\cite{xu2025qwen3} further pushes these boundaries by introducing a "multimodal thinking mode" and significantly reducing latency for extended inputs. Our work is positioned at this frontier, aiming to systematically evaluate and enhance MLLM capabilities in these complex, omnimodal scenarios.

\subsection{MLLM for Video Captioning}
The integration of Large Language Models (LLMs) has fundamentally reshaped video captioning, evolving from template-based outputs to context-aware narratives. Early VideoLLMs\cite{bai2025qwen3,zhang2025videollama,chen2026echoefficientchestxray,11551836,du2026unsupervised,du2026pansharpening,li2026unifusion,jiang2026imagine,jiang2025vknowu,liu2024tempcompass,zhang2025gapt,zhao2024toward,zhao2026spiking,zhou2026learning,zhao2025zero,hu2025exploiting,xie2024addsr,zhao2025spectral,zhao2024cycle,dong2025mamba,zhao2025multi,zhao2025learning,zhou2026more,zhang2025u} typically coupled pre-trained visual encoders with language backbones to capture dynamic transitions. To enhance granularity, recent models like OwlCap\cite{zhong2025owlcap} and the Tarsier series\cite{wang2024tarsier} leveraged large-scale SFT datasets to balance high-level motion semantics with fine-grained static details. However, these vision-centric models remain "deaf" to the audio modality, limiting their interpretation of real-world videos where acoustic cues are indispensable.
To overcome this, the field has pivoted toward omnimodal architectures \cite{xu2025qwen3,fu2024vita,zhan2024anygpt,lu2024unified,nguyen2025see,su2023pandagpt,zhang2025stream,zhao2026luve}. Proprietary models like Gemini 3 Pro\cite{gemini} and open-source breakthroughs such as Qwen3-Omni\cite{xu2025qwen3}] treat vision and audio as synchronized streams within a unified transformer space. Despite their general reasoning prowess, they are not explicitly optimized for the dense, temporally-aligned captioning essential for high-fidelity video understanding. While concurrent efforts like Video-SALMONN-2\cite{tang2025video} and UGC-VideoCaptioner\cite{wu2025ugc} employ intensive DPO or focus on specific content types, achieving precise temporal alignment between transient acoustic events and visual frames remains an open challenge. In this work, we aim to enhance temporal and cross-modal alignment  for joint audiovisual video captioning, ensuring that visual dynamics and acoustic semantics are harmoniously integrated into high-quality descriptions.
\section{TCA-Captioner}

\subsection{Challenge for Audiovisual Video Captioning}
Traditional video captioning models primarily follow a vision-centric paradigm, treat-ing audio as a redundant or auxiliary signal\cite{abdar2024review,zhou2024streaming,lin2022swinbert,chen2024sharegpt4video,chen2023diffusion,chen2025ragd,fan2025instancecap}. However, real-world audiovisual content is defined by high-density interactions where semantics are deeply entangled across modalities. Despite recent progress in joint audiovisual captioning, two primary bottlenecks remain largely overlooked: (1) \textbf{Audio-Visual Cross-Modal Binding}, the challenge of accurately associating a transient acoustic event (e.g., a specific utterance or a physical impact) with its corresponding visual source in a complex scene; and (2) \textbf{Audio-Visual Temporal Coherence}, the requirement to preserve the precise chronological and causal flow between auditory and visual streams. The neglect of these factors leads to "modality detachment," where captions may contain correct individual elements but fail to capture their synchronized relational logic.
\subsection{Observer-Checker-Corrector Framework}
\begin{figure}[t]
	\centering
	\includegraphics[width=1\linewidth]{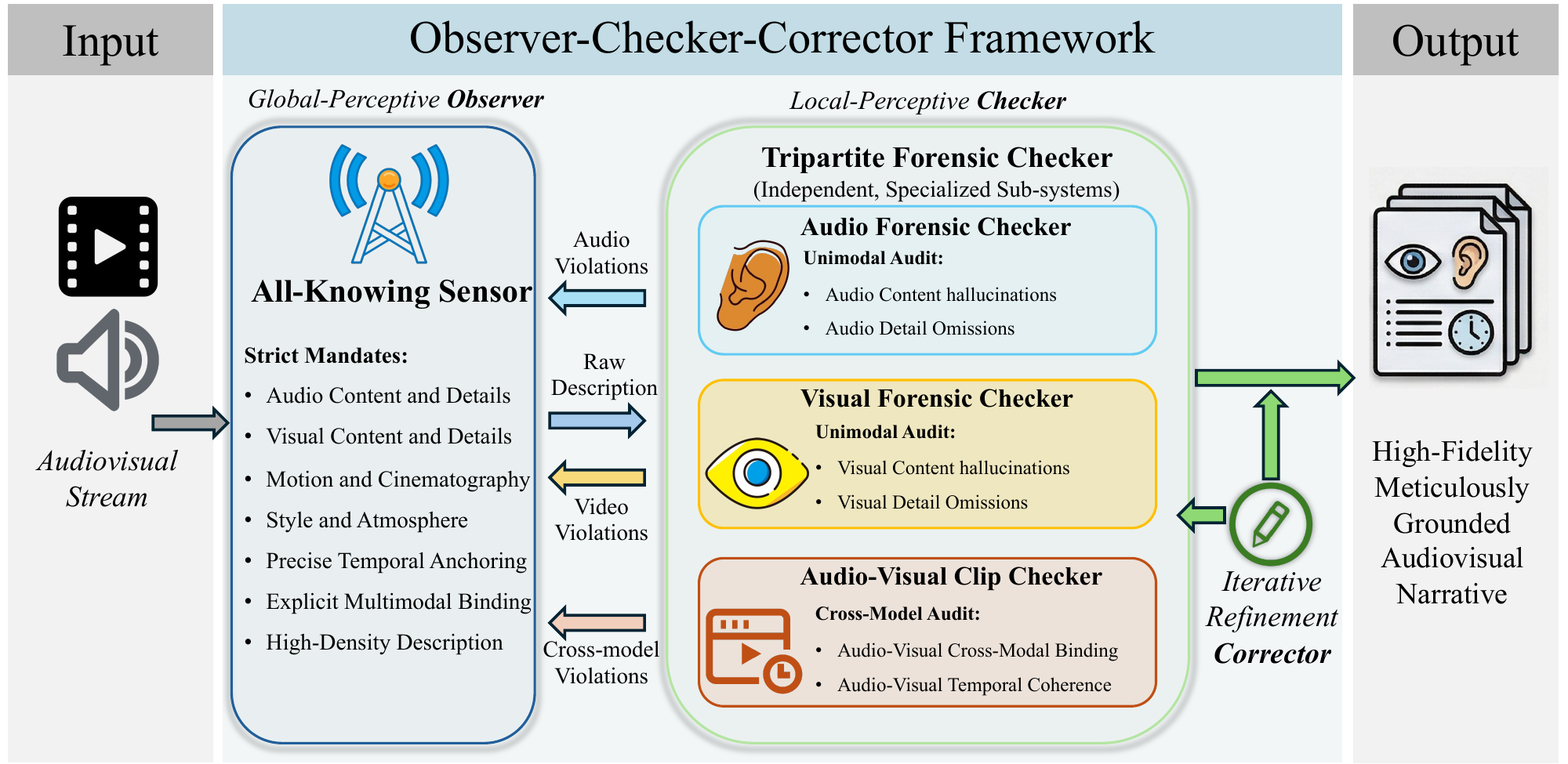}
 \vspace{-0.2cm}
	\caption{ The proposed Observer-Checker-Corrector (OCC) framework, an iterative data synthesis pipeline designed to generate high-quality audiovisual video captions. The framework consists of three collaborative modules: a Global-Perceptive Observer for comprehensive understanding, a Local-Perceptive Checker for detecting unimodal factual inaccuracies and cross-modal misalignments, and an Iterative Refinement Corrector for precision error correction and narrative enhancement. } 
	\vspace{-0.2cm}
	\label{fig:occ}
\end{figure}

To address the aforementioned challenges, we propose the Observer-Checker-Corrector (OCC) framework, an iterative data synthesis pipeline designed to generate high-fidelity, meticulously grounded audiovisual narratives. Unlike previous single-pass annotation, OCC employs a "divide-and-conquer" strategy through three collaborative modules, as shown in Fig. \ref{fig:occ}.

%\subsubsection{Global-Perceptive Observer}
\noindent{\textbf{Global-Perceptive Observer}}
The Observer serves as the primary perception engine, functioning as an all-knowing sensor that performs a global analysis of the audiovisual stream. Its objective is to produce a raw, high-density description.% that prioritizes objective reporting over semantic summarization.
The Observer is constrained by a strict mandate: 
(1) Audio Content and Details: Identify all sound events (speech, foley, ambient noise, background music) and their physical characteristics (loudness, frequency modulation, and timbre). 
(2) Visual Content and Details: Describe the scene, entities (appearance, actions), objects, lighting, and color palettes. 
(3) Motion and Cinematography: Catalog all subject movements (e.g., trajectories, gestures) and cinematographic techniques, including camera motion (pan, tilt, zoom) and shot transitions (cuts, fades).
(4) Aesthetic Style and Atmosphere: Characterize the overall stylistic tone (e.g., cinematic, documentary, handheld) and the auditory atmosphere (e.g., tense, reverberant, industrial).
(5) Precise Temporal Anchoring: All events must be localized with sub-second timestamps (e.g., [01.2s - 01.5s]) to ensure strict alignment with the raw signal.
(6) Explicit Multimodal Binding: Avoiding vague pronouns in favor of entity-specific attribution (e.g., grounding a specific segment of speech to its corresponding speaker among multiple candidates in a scene).
(7) High-Density Description: Prioritize a comprehensive inventory of atomic facts over abstract summaries. %the output should be a dense, structured log of every perceptible change in the audiovisual environment. 
We employ Gemini-3-Pro, a formidable MLLM, as the Observer to navigate the inherent intricacies of global audiovisual analysis.%We employ Gemini-3-Pro as the Observer.
%It must distinguish between diegetic (on-screen) and non-diegetic (off-screen) sounds.
%Every visual element must be described with its spatial coordinates and state changes.
%, to preserve the structural rhythm of the video
%to provide a contextual baseline for high-level semantic reasoning
%(i) Precise Temporal Anchoring: all events must be localized with sub-second timestamps (e.g., [01.2s - 01.5s]); (ii) Explicit Multimodal Binding: avoiding vague pronouns in favor of entity-specific attribution (e.g., linking specific lip movements to a corresponding vocal track); and (iii) \textbf{Factual Transparency}: reporting any perceived audiovisual lag or mismatch exactly as captured, without attempting to "fix" the scene's logic.

\noindent{\textbf{Local-Perceptive Checker}}
To guarantee factual integrity, we introduce a Tripartite Forensic Checker consisting of three independent, specialized sub-systems. This decoupled verification ensures that unimodal errors do not propagate into cross-modal misalignments.
\begin{itemize}
\item \textbf{Audio Forensic Checker}: This module performs an independent audit of the audio track to verify auditory hallucinations and identify omissions in granular sound details. Focusing on Verbal Fidelity and Linguistic Integrity, it strictly penalizes phonetic discrepancies and flags "phantom sounds" that lack an acoustic basis. Beyond content validation, the checker scrutinizes fine-grained attributes, such as timbre characteristics, ambient interference, and rhythmic prosody, while rigorously ensuring the temporal coherence of acoustic events to maintain precise onset and offset alignment. We employ Gemini-3-Pro as the audio checker to ensure the factual integrity and perceptual accuracy of the auditory information.
\item \textbf{Visual Forensic Checker}: Acting on silent video tracks, this module is engineered to audit visual hallucinations and recover omitted fine-grained details. It prioritizes the validation of entity dynamics and spatio-temporal logic, meticulously scrutinizing micro-expressions and motion trajectories to ensure the narrative aligns with the physical causality of the visual scene. Furthermore, it supplements any overlooked granular nuances while rigorously enforcing the temporal coherence of visual events, ensuring that the chronological flow of actions remains logically sound and consistent. We observe that Gemini-3-Pro introduces a significant amount of hallucinations when processing silent video streams. Therefore, we employ the powerful silent video understanding model, Doubao-1.8, as the Visual Checker, which possesses more accurate and fine-grained visual understanding capabilities.
\item \textbf{Audio-Visual Clip Checker}: This is the core module for evaluating audio-visual cross-modal binding and temporal coherence. It distinguishes between diegetic sounds (on-screen), off-screen sounds, and non-diegetic voice-overs. It specifically targets "lip-sync" discrepancies and ensures that the narrative preserves the chronological sequence of cross-modal interactions. To achieve this, the module employs a sliding-window strategy, partitioning the video into 2-second clips that are processed iteratively to generate detailed verification logs. Notably, each input segment is augmented with its sequence index and corresponding timestamps, facilitating superior calibration of temporal coherence. Given the intensive computational demands of this granular analysis, we utilize Gemini-3-Flash as the underlying engine for the Clip Checker, balancing high-fidelity reasoning with operational efficiency.
\end{itemize}

%\subsubsection{Iterative Refinement Corrector}
\noindent{\textbf{Iterative Refinement Corrector}}
The Corrector acts as the final synthesis agent. It receives the initial description from the Observer and the comprehensive audit reports (violation lists) from the tripartite Checkers. Its task is to perform an optimal fusion: rectifying hallucinations identified by the unimodal checkers and resolving alignment conflicts flagged by the audiovisual checker. Through this iterative feedback loop, the Corrector produces a final caption that is both semantically rich and structurally aligned with the raw sensor data. We employ Gemini-3-Pro as the refinement corrector. 

\noindent{\textbf{Token Cost}}
As detailed in Table \ref{tab:tokens}, processing a 10-second video consumes approximately 22,000 tokens (both input and output) across the OCC pipeline. The A/V-Checker accounts for the majority of this cost due to its iterative processing of the entire audio-visual sequence.

\subsection{High-Density Human Interaction Dataset}

\begin{table}[t]
\centering
% \vspace{-0.7 cm}
\caption{Token Consumption per 10s Video.}
\label{tab:tokens}
\resizebox{\linewidth}{!}{%
\begin{tabular}{cccccc}
\toprule
 \textbf{Observer} & \textbf{A-Checker} & \textbf{V-Checker} & \textbf{A/V-Clip-Checker} & \textbf{Corrector} & \textbf{Total} \\
\midrule
3,500 & 1,535 & 3,874 & 8,705 & 4,335 & \textbf{21,949} \\
\bottomrule
\end{tabular}}
% \vspace{-0.7 cm}
    \vspace{-0.2cm}
\end{table}
% Existing video datasets \cite{tiktok_10m_2025,Farré2024FineVideo,shang2025large} often feature sparse audiovisual events, making them insufficient for training models on complex alignment tasks. To fill this gap, we curate the High-Density Human Interaction (HHI) Dataset, comprising 3,500 meticulously selected video clips sourced from YouTube and cinematic content.
% The HDI dataset is specifically filtered for "hard cases": multi-speaker dialogues in noisy environments, rapid physical actions coupled with impact sounds, and subtle cross-modal causal chains. Each clip is processed through our OCC pipeline, yielding high-fidelity annotations. This corpus provides the necessary supervision for the TCA-Captioner to transition from coarse scene-level understanding to fine-grained, synchronized audiovisual perception. 

Existing video datasets\cite{tiktok_10m_2025,Farré2024FineVideo,shang2025large} often feature sparse audiovisual events, rendering them insufficient for training models on complex alignment tasks. To address this gap, we curate the High-Density Human Interaction (HDI) Dataset, comprising 3,500 meticulously selected clips sourced from YouTube and cinematic content. The HDI dataset is specifically filtered for "hard cases," such as multi-speaker dialogues in noisy environments, rapid physical actions coupled with impact sounds, and subtle cross-modal causal chains. Each clip is processed through our full OCC pipeline, yielding high-fidelity annotations that provide the necessary supervision for the TCA-Captioner to transition from coarse scene-level understanding to fine-grained, synchronized audiovisual perception.
Furthermore, to enhance the diversity of the training distribution, we incorporate an additional 5,000 samples from FineVideo\cite{Farré2024FineVideo} and 7,000 samples from TikTok-10M\cite{tiktok_10m_2025}. Given the extended duration and relatively straightforward content of these videos, we streamlined the annotation process by bypassing the Clip Checker module within the OCC framework. Leveraging this comprehensive multi-source corpus, we perform LoRA fine-tuning on the Qwen3-Omni\cite{xu2025qwen3} backbone to develop our final TCA-Captioner.

% --- Section 5: TCA-Bench ---
\section{TCA-Bench}
\label{sec:tca_bench}

\subsection{Motivation}
While recent audiovisual captioning benchmarks have advanced the measurement of detailed perception, they fundamentally assess only the \textit{static presence} of individual details, not the dynamic interplay between modalities. Video-SALMONN-2 \cite{tang2025video} decomposes videos into flat atomic events to compute missing and hallucination rates; UGC-VideoCap \cite{wu2025ugc} assigns a single holistic LLM score across entangled dimensions; Omni-Cloze \cite{ma2025omni} transforms evaluation into cloze-style multiple-choice questions to penalize hallucinations on isolated attributes. 
A common structural limitation underlies all three: they treat multimodal content as an unordered bag of independent facts, ignoring the relational structure that connects modalities. Consequently, none of these paradigms can systematically evaluate whether a model correctly \textit{binds} a sound to its visual source or preserves the \textit{temporal ordering} of cross-modal events, two capabilities we identify as the primary failure modes of state-of-the-art models. Table~\ref{tab:bench_comparison} contrasts these benchmarks with TCA-Bench. To bridge this gap, TCA-Bench introduces a \textbf{Decoupled Evaluation Protocol}, specifically designed to disentangle foundational unimodal perception from higher-order relational reasoning.
% In practice, even when a model successfully recalls every individual visual and audio element, it frequently suffers from modality detachment (misattributing a sound to the wrong visual source) and temporal incoherence (inverting the order of cross-modal events). TCA-Bench is designed to isolate and quantify exactly these failures through a \textbf{Decoupled Evaluation Protocol} that disentangles foundational uni-modal perception from higher-order relational reasoning.

% Recent audiovisual captioning benchmarks, such as Video-SALMONN-2 \cite{tang2025video}, UGC-VideoCap \cite{wu2025ugc}, and Omni-Cloze \cite{ma2025omni}, primarily measure the static presence of isolated details or assign holistic scores to entangled dimensions. 
% A common structural limitation persists: they treat multimodal content as an unordered bag of independent facts, failing to account for the relational structure between modalities. Consequently, these paradigms cannot systematically evaluate modality binding (linking sound to its visual source) or temporal ordering—two critical dimensions where state-of-the-art models frequently falter. To bridge this gap, TCA-Bench introduces a Decoupled Evaluation Protocol, specifically designed to disentangle foundational unimodal perception from higher-order relational reasoning (see Table~\ref{tab:bench_comparison}).

% --- Table: Caption ABOVE the table ---
\begin{table}[t]
\caption{Comparison of audiovisual video captioning benchmarks. The upper block compares evaluation capabilities; the lower block compares dataset properties. $\checkmark$ = explicitly supported; $\times$ = not supported; (P) = partial.}
\label{tab:bench_comparison}
\centering
\resizebox{\textwidth}{!}{
\begin{tabular}{lcccc}
\toprule
Dimension & video-SALMONN-2 & UGC-VideoCap & Omni-Cloze & \textbf{TCA-Bench} \\ \midrule
A-V Binding Evaluation & $\times$ & $\times$ & $\times$ & $\checkmark$ \\
A-V Temporal Evaluation & $\times$ & $\times$ & $\times$ & $\checkmark$ \\
Evaluation Paradigm & Event Matching & Holistic Scoring & Cloze-style MC & Decoupled Checklist \\
Independent Uni-modal Scoring & $\times$ & $\times$ & (P) & $\checkmark$ \\
Structured Cross-Modal GT & $\times$ & $\times$ & $\times$ & $\checkmark$ \\ \midrule
Number of Videos & 483 & 1,000 & 2,340 & 459 \\
A-V Interaction Density & Low & Low & Low & High \\
\bottomrule
\end{tabular}
}
\end{table}

\subsection{Benchmark Construction}

\begin{figure}[t]
	\centering
	\includegraphics[width=1\linewidth]{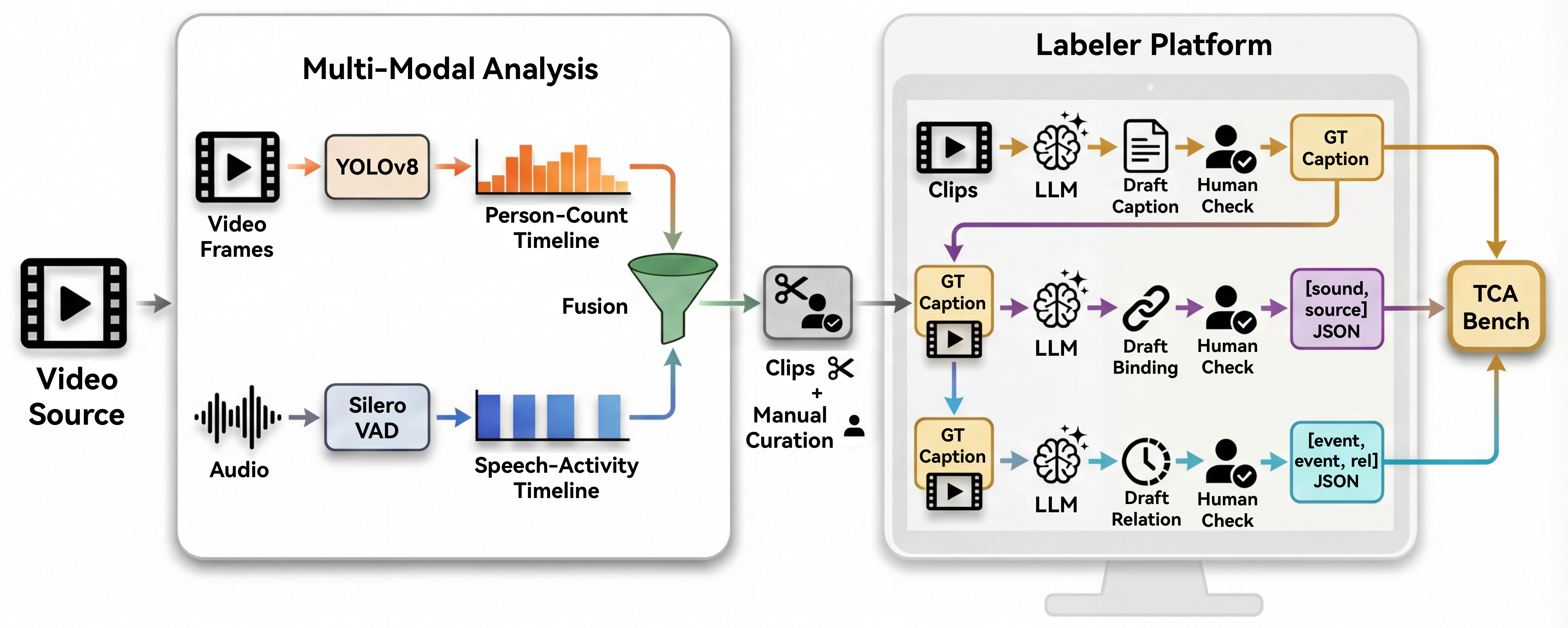}
 \vspace{-0.2cm}
	\caption{ TCA-Bench data curation and annotation pipeline. \textit{Stage I}: source videos undergo parallel visual (YOLOv8 \cite{varghese2024yolov8}) and auditory (Silero VAD \cite{SileroVAD}) analysis; timelines are fused under a joint co-occurrence constraint to extract segments. \textit{Stage II}: each clip receives three ground-truth layers via LLM draft $\rightarrow$ human validation. } 
	\vspace{-0.2cm}
	\label{fig:pipeline}
\end{figure}

\paragraph{Data curation.} TCA-Bench targets videos with dense audio-visual interplay, specifically multi-person dialogues coupled with complex physical actions (e.g., variety shows, group interactions)---sourced from YouTube. Such scenes produce frequent, entangled cross-modal events that stress-test audiovisual binding and temporal relational reasoning. We focus on short clips (5--30 seconds), as these densely interleaved segments already pose substantial challenges for current models. As illustrated in Fig.~\ref{fig:pipeline}, the construction pipeline proceeds in two macro-stages: (I) automated clip curation via multi-modal temporal analysis, and (II) structured annotation on a purpose-built labeler platform.

% --- Figure: Caption BELOW the figure ---

In Stage I, two parallel analysis streams operate on each source video: a visual stream uses YOLOv8 for person detection to produce a per-frame person-count timeline, while an auditory stream applies Silero VAD to generate speech-activity intervals. The two timelines are fused under a joint co-occurrence constraint, retaining only intervals where multiple persons are visually present \textit{and} speech is simultaneously active, followed by quality filtering on person-detection ratio and speech overlap. Human annotators then perform a final curation pass to ensure sufficient density and diversity of cross-modal interactions. In Stage II, each curated clip enters a purpose-built web-based labeler platform where all three annotation layers are produced through a unified \textit{LLM-draft-then-human-validation} paradigm: a multimodal LLM first generates an initial dense caption, which human annotators correct and finalize; the verified caption then serves as context for the LLM to draft binding pairs and temporal triples, each again undergoing human validation.

\paragraph{Structured annotation.} Unlike prior benchmarks that rely solely on flat textual ground truths, TCA-Bench formalizes cross-modal dependencies with three complementary annotation layers:
\begin{enumerate}
    \item \textbf{Ground-truth captions} that exhaustively describe visual details (subjects, actions, scenes) and auditory elements (speech, sounds).
    \item \textbf{Audio-Visual Binding Lists} --- structured JSON lists of \texttt{[sound, source]} pairs linking each acoustic event to categories: \textit{foreground character}, \textit{off-screen character}, or \textit{environment}.
    \item \textbf{Temporal Relational Lists} --- triples of \texttt{[first\_event, second\_event, relation]} encoding relation types: \texttt{A\_then\_V}, \texttt{V\_then\_A}, or \texttt{AV\_simultaneous}.
\end{enumerate}

\begin{figure}[t]
	\centering
	\includegraphics[width=1\linewidth]{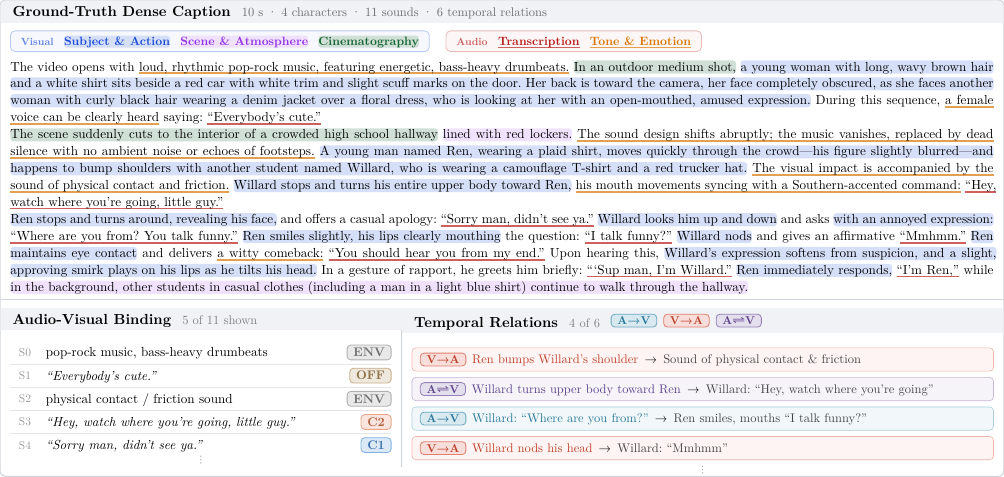}
 	\vspace{-0.2cm}
	\caption{TCA-Bench ground-truth annotation showcase. \textbf{Top:} a dense caption with inline highlights delineating five evaluation sub-dimensions grouped into two modality tracks (Visual: Subject \& Action, Scene \& Atmosphere, Cinematography; Audio: Transcription Accuracy, Tone \& Emotion). \textbf{Bottom-left:} audio-visual binding pairs mapping each acoustic event to its source—foreground character, off-screen character, or environment. \textbf{Bottom-right:} temporal relational triples encoding cross-modal sequential ($A{\rightarrow}V$, $V{\rightarrow}A$) and simultaneous ($A{\rightleftharpoons}V$) orderings. The clip is chosen for balanced coverage across all sub-dimensions, source categories, and relation types.} 
 	\vspace{-0.2cm}
	\label{fig:showcase}
\end{figure}

These three layers are explicitly aligned with the evaluation protocol: ground-truth captions support Base Perception scoring, while Binding Lists and Temporal Relational Lists serve as structured references for the cross-modal Binding Accuracy and Temporal $F_1$ metrics, respectively. Fig.~\ref{fig:showcase} illustrates the three annotation layers on a representative benchmark clip.

\subsection{Decoupled Evaluation Protocol}
Existing evaluation paradigms conflate uni-modal perception errors with cross-modal reasoning failures, making it impossible to diagnose \textit{where} a model fails. TCA-Bench addresses this with a two-tier \textbf{Decoupled Evaluation Protocol} using an independent LLM judge.

\paragraph{Base Perception (Uni-modal).} An LLM judge compares the candidate against the GT caption on a 0--10 scale:
\begin{itemize}
    \item \textbf{Audio:} \textit{Transcription Accuracy}---verbatim fidelity of speech content including meaning-critical words and fillers; \textit{Tone \& Emotion}---accuracy of the described emotional delivery style and non-speech audio such as music mood and ambient sounds.
    \item \textbf{Visual:} \textit{Subject \& Action}---correctness of primary subjects' appearance, quantity, spatial relationships, and sequential movements; \textit{Scene \& Atmosphere}---fidelity of the setting, background layout, spatiotemporal attributes, and visual mood; \textit{Cinematography}---accuracy of shot size, camera angles, and camera movement patterns.
\end{itemize}

\paragraph{Audiovisual Integration (Cross-Modal).} This track uses structured GT annotations for targeted checklist evaluation:
\begin{itemize}
    \item \textbf{Binding Accuracy:} For each GT binding pair, the judge performs a two-stage verification: first confirming that the acoustic event is \textit{present} in the candidate caption, then checking whether it is attributed to the \textit{correct source}. A sound is scored as \textit{correct} only when both stages pass. Crucially, this decoupled design ensures that transcription errors (e.g., mishearing speech content) do not penalize binding scores---such errors are isolated in the Base Perception tier. We report accuracy separately for \textit{Character} (foreground + off-screen) and \textit{Environment} sounds.
    \item \textbf{Temporal $F_1$:} For each triple in the GT temporal list, the judge assigns a three-way verdict: \textit{correct} (both events present with a consistent temporal link), \textit{incorrect} (both events present but the temporal relationship is clearly reversed or contradicted), or \textit{skipped} (one or both events absent). This three-way protocol decouples temporal \textit{accuracy} from event \textit{coverage}: omitting an event yields a \textit{skipped} verdict rather than a false temporal error. We compute:
\end{itemize}

\begin{equation}
\text{Prec} = \frac{|\text{correct}|}{|\text{correct}| + |\text{incorrect}|}, \quad \text{Cov} = \frac{|\text{correct}| + |\text{incorrect}|}{|\text{GT}|}
\end{equation}

\begin{equation}
F_1 = \frac{2 \cdot \text{Prec} \cdot \text{Cov}}{\text{Prec} + \text{Cov}}
\end{equation}

We report $F_1$ separately for \textit{Sequential} and \textit{Simultaneous} relations.

\section{Experiment}

\subsection{Caption Evaluation on Public Benchmarks}

\begin{table}[t]
    \centering
    \caption{Caption evaluation on the public benchmarks. ``A'' and ``V'' refer to the audio and visual modalities, respectively. The results presented above are reproduced using the official code. We strictly adhere to the evaluation protocol established by AVoCaDO to ensure a fair and consistent comparison.}	\vspace{-0.2cm}
    \label{tab:publicbench}
    \resizebox{\linewidth}{!}{
    \begin{tabular}{lcc|ccc|cccc}
    \toprule
        \multirow{2}{*}[-0.1cm]{\textbf{Model}} & \multirow{2}{*}[-0.1cm]{\textbf{Size}} & \multirow{2}{*}[-0.1cm]{\textbf{Modality}} & \multicolumn{3}{c|}{\textbf{video-SALMONN-2}} & \multicolumn{4}{c}{\textbf{UGC-VideoCap}} \\
        \cmidrule(lr){4-6} \cmidrule(lr){7-10}
        ~ & ~ & ~ & Miss~$\downarrow$ & Hall.~$\downarrow$ & Total~$\downarrow$ & Audio~$\uparrow$ & Visual~$\uparrow$ & Detail~$\uparrow$ & Avg.~$\uparrow$ \\
        \midrule
        \textcolor{gray!100}{Gemini-2.5-Pro} & \textcolor{gray!100}{-} & \textcolor{gray!100}{A + V} & \textcolor{gray!100}{18.1} & \textcolor{gray!100}{13.3} & \textcolor{gray!100}{31.3} & \textcolor{gray!100}{69.5} & \textcolor{gray!100}{74.7} & \textcolor{gray!100}{73.7} & \textcolor{gray!100}{72.6} \\
        \textcolor{gray!100}{Gemini-2.5-Flash} & \textcolor{gray!100}{-} & \textcolor{gray!100}{A + V} & \textcolor{gray!100}{19.3} & \textcolor{gray!100}{13.9} & \textcolor{gray!100}{33.3} & \textcolor{gray!100}{69.1} & \textcolor{gray!100}{75.8} & \textcolor{gray!100}{74.0} & \textcolor{gray!100}{73.0} \\
        \textcolor{gray!100}{Gemini-3.0-Pro} & \textcolor{gray!100}{-} & \textcolor{gray!100}{A + V} & \textcolor{gray!100}{18.9} & \textcolor{gray!100}{14.8} & \textcolor{gray!100}{33.7} & \textcolor{gray!100}{71.3} & \textcolor{gray!100}{76.7} & \textcolor{gray!100}{77.8} & \textcolor{gray!100}{75.3} \\
        % \textcolor{gray!100}{Gemini-3.0-Flash} & \textcolor{gray!100}{-} & \textcolor{gray!100}{A + V} & \textcolor{gray!100}{16.6} & \textcolor{gray!100}{10.4} & \textcolor{gray!100}{27.0} & \textcolor{gray!100}{73.0} & \textcolor{gray!100}{78.8} & \textcolor{gray!100}{77.9} & \textcolor{gray!100}{76.6} \\
        
        \midrule
        % InternVL3.5 & 8B & V & 53.8 & 25.5 & 79.4 & 47.9 & 64.8 & 59.5 & 57.4 \\
        % Qwen2.5-VL & 7B & V & 40.5 & 17.0 & 57.5 & 46.6 & 69.1 & 62.3 & 59.3 \\
        % \midrule
        HumanOmniV2 & 7B & A + V & 49.2 & \textbf{12.3} & 61.6 & 45.6 & 66.3 & 59.5 & 57.1 \\
        ARC-Hunyuan-Video & 7B & A + V & 45.7 & 12.5 & 58.2 & 52.7 & 56.0 & 55.8 & 54.8 \\
        MiniCPM-o-2.6 & 8B & A + V & 42.2 & 14.3 & 56.5 & 38.6 & 68.5 & 57.7 & 54.9 \\
        Qwen2.5-Omni & 7B & A + V & 41.7 & 15.4 & 57.1 & 46.9 & 66.1 & 60.0 & 57.7 \\
        Qwen3-Omni & 30B-A3B & A + V & 32.0 & 13.6 & 45.6 & 67.5 & 74.8 & 72.3 & 71.5 \\
        video-SALMONN-2 & 7B & A + V & 21.2 & 17.6 & 38.8 & 61.8 & 71.4 & 68.5 & 67.2 \\
        UGC-Captioner & 3B & A + V & 31.6 & 17.0 & 48.6 & 61.4 & 58.4 & 57.5 & 59.1 \\
        % \textcolor{gray!100}{Qwen3-Omni-Captioner} & \textcolor{gray!100}{30B-A3B} & \textcolor{gray!100}{A + V} & \textcolor{gray!100}{31.0} & \textcolor{gray!100}{16.6} & \textcolor{gray!100}{47.6} & \textcolor{gray!100}{69.0} & \textcolor{gray!100}{75.5} & \textcolor{gray!100}{72.3} & \textcolor{gray!100}{72.5} \\
        AVoCaDO  & 7B & A + V & \underline{21.1} & 16.2 & \underline{37.3} & \underline{73.0} & \underline{74.6} & \underline{71.8} & \underline{73.2} \\
        \midrule
        TCA-Captioner & 30B-A3B & A + V & \textbf{18.6} & \underline{12.9} & \textbf{31.5} & \textbf{73.8} & \textbf{76.0} & \textbf{72.9} & \textbf{74.2} \\

    \bottomrule
    \end{tabular}
    	
    }
    \vspace{-0.2cm}
\end{table}

% We first benchmark the audiovisual captioning proficiency of TCA-Captioner across two public rigorous evaluation platforms: the Video-SALMONN-2 testset and the UGC-VideoCap benchmark. As detailed in Table \ref{tab:publicbench}, TCA-Captioner consistently establishes a new performance ceiling, eclipsing all existing open-source baselines, including the most recent AVoCaDO.
% While specific models like HumanOmniV2 may report a marginally lower Hallucination rate on the Video-SALMONN-2 set, a closer inspection reveals this is a byproduct of their conservative, overly concise narrative style. Such brevity inherently limits their descriptive utility, as evidenced by their elevated Miss rates and diminished scores on the more detail-oriented UGC-VideoCap benchmark. In contrast, TCA-Captioner achieves a superior equilibrium between descriptive density and factual precision. By effectively capturing nuanced interactions without sacrificing accuracy, our model significantly outperforms its counterparts in the Total metric of Video-SALMONN-2 and across the average performance indices of UGC-VideoCap.
% Furthermore, when positioned against the latest large-scale MoE architectures like Qwen3-Omni, TCA-Captioner maintains a distinct competitive edge. Most notably, our model demonstrates remarkable robustness by surpassing the Gemini 3.0 series on the UGC-VideoCap benchmark. This superior performance validates the efficacy of our temporal and cross-modal alignment strategies, underscoring TCA-Captioner's leading capability in synthesizing high-fidelity audiovisual narratives.

We first evaluate the audiovisual captioning proficiency of TCA-Captioner across two  public benchmarks: the Video-SALMONN-2 \cite{tang2025video} and the UGC-VideoCap \cite{wu2025ugc} benchmark. To ensure a fair and consistent comparison, we strictly adhere to the evaluation protocol established by AVoCaDO\cite{chen2025avocado}, employing GPT-4.1 as the judge model for Video-SALMONN-2 and GPT-4o for UGC-VideoCap. Our model is benchmarked against a comprehensive suite of representative baselines, including prominent open-source models, such as HumanOmniV2\cite{yang2025humanomniv2}, ARC-Hunyuan-Video\cite{ge2025arc}, MiniCPM-o-2.6\cite{yao2024minicpm}, Qwen2.5-Omni\cite{xu2025qwen25omnitechnicalreport}, Qwen3-Omni\cite{xu2025qwen3}, video-SALMONN-2\cite{tang2025video}, UGC-Captioner\cite{wu2025ugc}, and the latest AVoCaDO\cite{chen2025avocado}, as well as the powerful closed-source Gemini series \cite{gemini}.
As illustrated in Table \ref{tab:publicbench}, TCA-Captioner achieves state-of-the-art (SOTA) performance among all open-source models and demonstrates formidable competitiveness even when positioned against top-tier closed-source systems. 
% While certain models like HumanOmniV2 report a marginally lower Hallucination (Hall.) rate on Video-SALMONN-2, a closer inspection reveals that this is a byproduct of an overly conservative and concise narrative style. Such brevity inherently limits descriptive utility, as evidenced by their diminished scores on the detail-oriented UGC-VideoCap benchmark. In contrast, TCA-Captioner strikes a superior equilibrium between descriptive density and factual precision. 
% By effectively capturing nuanced interactions without sacrificing accuracy, our model significantly outperforms its counterparts in the Total metric of Video-SALMONN-2 and the average performance indices of UGC-VideoCap, eclipsing all open-source competitors including AVoCaDO.
%Notably, TCA-Captioner exhibits remarkable robustness: on the UGC-VideoCap benchmark, its average score (74.2) not only vastly exceeds other open-source models but also surpasses the Gemini-2.5 series while remaining on par with the cutting-edge Gemini-3.0 series. This exceptional performance validates the efficacy of our proposed strategies and underscores TCA-Captioner’s leading capability in synthesizing high-fidelity audiovisual narratives.

\subsection{Caption Evaluation on Our TCA-Bench}

\begin{table}[t]
    \centering
    \caption{Caption evaluation on our TCA-Bench.  We employ GPT-4.1 as the judge model. Note that all metric scores have been normalized to a scale of 0–100.}
    	\vspace{-0.2cm}
    \label{tab:ourbench}
    \resizebox{\linewidth}{!}{
    \begin{tabular}{l|ccc|cccc|ccc|ccc}
    \toprule
        \multirow{2}{*}[-0.1cm]{\textbf{Model}}  & \multicolumn{3}{c|}{\textbf{Audio}} & \multicolumn{4}{c|}{\textbf{Visual}} & \multicolumn{3}{c|}{\textbf{AV Binding}} & \multicolumn{3}{c}{\textbf{AV Temporal}}\\
        \cmidrule(lr){2-4} \cmidrule(lr){5-8} \cmidrule(lr){9-11} \cmidrule(lr){12-14}
        ~  & Trans.~$\uparrow$ & Ton.~$\uparrow$ & Avg.~$\uparrow$ & Sub.~$\uparrow$ & Sce.~$\uparrow$ & Cine.~$\uparrow$ & Avg.~$\uparrow$ & Char.~$\uparrow$ & Envi.~$\uparrow$ & Total~$\uparrow$ & Seq.~$\uparrow$ & Sim.~$\uparrow$ & Total~$\uparrow$\\
        \midrule

       %  Gemini-2.5-Pro  & 64.2  & 66.3 & 65.3 & 66.0 & 69.5 & 63.1 & 66.2  & 82.6 & 47.6 & 74.3 & 78.5 &76.9 &78.0\\
       %  Gemini-2.5-Flash  & 62.2 & 64.6 & 63.4 & 66.0 &69.6  & 67.2 & 67.6  & 77.3 & 41.1 & 68.6 & 73.3 & 71.6&72.7 \\
       %  Gemini-3.0-Pro  & 71.5 & 68.9 & 70.2 & 61.5 & 62.8 & 58.5 &60.9  & 87.3 & 42.8 & 76.8 & 80.9 & 78.3 & 80.0 \\
       % Gemini-3.0-Flash &  61.7 & 59.6  & 60.7 & 60.0 & 63.2 & 50.1 & 57.8&81.3&34.3&70.1&72.4&71.4&72.0 \\
       \textcolor{gray!100}{Gemini-2.5-Pro}  & \textcolor{gray!100}{64.2}  & \textcolor{gray!100}{66.3} & \textcolor{gray!100}{65.3} & \textcolor{gray!100}{66.0} & \textcolor{gray!100}{69.5} & \textcolor{gray!100}{63.1} & \textcolor{gray!100}{66.2}  & \textcolor{gray!100}{82.6} & \textcolor{gray!100}{47.6} & \textcolor{gray!100}{74.3} & \textcolor{gray!100}{78.5} &\textcolor{gray!100}{76.9} &\textcolor{gray!100}{78.0}\\
\textcolor{gray!100}{Gemini-2.5-Flash}  & \textcolor{gray!100}{62.2} & \textcolor{gray!100}{64.6} & \textcolor{gray!100}{63.4} & \textcolor{gray!100}{66.0} &\textcolor{gray!100}{69.6}  & \textcolor{gray!100}{67.2} & \textcolor{gray!100}{67.6}  & \textcolor{gray!100}{77.3} & \textcolor{gray!100}{41.1} & \textcolor{gray!100}{68.6} & \textcolor{gray!100}{73.3} & \textcolor{gray!100}{71.6}&\textcolor{gray!100}{72.7} \\
\textcolor{gray!100}{Gemini-3.0-Pro}  & \textcolor{gray!100}{71.5} & \textcolor{gray!100}{68.9} & \textcolor{gray!100}{70.2} & \textcolor{gray!100}{61.5} & \textcolor{gray!100}{62.8} & \textcolor{gray!100}{58.5} &\textcolor{gray!100}{60.9}  & \textcolor{gray!100}{87.3} & \textcolor{gray!100}{42.8} & \textcolor{gray!100}{76.8} & \textcolor{gray!100}{80.9} & \textcolor{gray!100}{78.3} & \textcolor{gray!100}{80.0} \\
\textcolor{gray!100}{Gemini-3.0-Flash} &  \textcolor{gray!100}{61.7} & \textcolor{gray!100}{59.6}  & \textcolor{gray!100}{60.7} & \textcolor{gray!100}{60.0} & \textcolor{gray!100}{63.2} & \textcolor{gray!100}{50.1} & \textcolor{gray!100}{57.8}&\textcolor{gray!100}{81.3}&\textcolor{gray!100}{34.3}&\textcolor{gray!100}{70.1}&\textcolor{gray!100}{72.4}&\textcolor{gray!100}{71.4}&\textcolor{gray!100}{72.0} \\
        \midrule
        Qwen2.5-Omni-3B  & 29.3 & 32.3 & 30.8 & 46.9 & 52.1 & 29.1 & 42.7  & 41.6 & 17.2 & 35.8 & 33.3 &38.0 & 35.0\\
        Qwen2.5-Omni-7B  & 33.6 &37.8 & 35.7 & 51.1 & 54.6 &29.9 & 45.2 & 39.9  &18.3  &34.8  &37.5  &43.3  &39.7\\
        Qwen3-Omni & 35.2  & 42.8 & 39.0 & 53.3 & 61.7 & 39.2 & 51.4 & 48.8 & 22.2 & 42.5 & 46.7 & 49.9 &47.9\\
        UGC-Captioner & 39.8 & 48.2 & 44.0 & 48.7  & 58.8 & 35.0 & 47.5 & 51.5 & 22.3 & 44.6 & 46.1 & 51.0 &47.9 \\
        AVoCaDO   & \underline{59.2} &  \textbf{62.3} & \underline{60.8} & \underline{59.3} & \underline{63.9} &  \textbf{65.2} & \underline{62.8} & \underline{76.7} & \underline{36.8} &\underline{67.2}  &  \underline{68.8}&\underline{67.3}&\underline{68.3} \\
        \midrule
        TCA-Captioner  & \textbf{61.6} &   \underline{60.8} & \textbf{61.2} & \textbf{59.6} & \textbf{68.9} & \underline{62.2}  & \textbf{63.6} & \textbf{82.1} & \textbf{44.6} & \textbf{73.2} & \textbf{76.7} &\textbf{77.1}&\textbf{76.9}\\

    \bottomrule
    \end{tabular}	
    }
    \vspace{-0.2cm}
\end{table}

We conduct extensive experiments on our proposed TCA-Bench to evaluate the multi-modal proficiency of TCA-Captioner against a broad spectrum of models, including the closed-source Gemini series (v2.5 and v3.0) and various leading open-source MLLMs. The performance is measured across four primary dimensions: Audio (Transcription, Tone $\&$ Emotion), Visual (Subject $\&$ Action, Scene $\&$  Atmosphere, Cinematography), Audiovisual (AV) Binding (Character, Environment), and AV Temporal (Sequential, Simultaneous). The experimental results in Table \ref{tab:ourbench} reveal several key insights:

\noindent{\textbf{Gap between Open and Closed-Source Models}}: A significant performance disparity exists between existing open-source models and the closed-source Gemini series. This gap is particularly pronounced in the AV Binding and AV Temporal tracks, where open-source models struggle to accurately associate acoustic events with their corresponding visual sources or maintain the logical chronological order of cross-modal interactions.

\noindent{\textbf{Evolution of Gemini Series}}: Gemini-3.0-Pro demonstrates substantial improvements in Audio (70.2 avg.), AV Binding (76.8 total) and AV Temporal (80.0 total) compared to its predecessor, Gemini-2.5-Pro (65.3, 74.3 and 78.0, respectively). However, a performance degradation is observed in the Visual dimension, where its average score drops from 66.2 to 60.9, suggesting a trade-off in its visual perception tuning.

\noindent{\textbf{Superiority of TCA-Captioner}}: 
Our model consistently outperforms all existing open-source models across both the average and total metrics within the Audio, Visual, AV Binding, and AV Temporal dimensions. Notably, in the AV Binding (73.2) and AV Temporal (76.9) categories, TCA-Captioner exhibits a profound understanding of cross-modal interactions, nearly doubling the performance of the baseline Qwen3-Omni.
% Our model consistently outperforms all open-source models by a wide margin. Notably, in the AV Binding (73.2) and AV Temporal (76.9) categories, TCA-Captioner exhibits a profound understanding of cross-modal interactions, nearly doubling the performance of the baseline Qwen3-Omni.

\noindent{\textbf{Competitive Edge against Closed-Source Models}}: Compared to the Gemini family, TCA-Captioner surpasses Gemini-3.0-Flash in every evaluated dimension. Furthermore, our model achieves a superior Video average score (63.6) compared to Gemini-3.0-Pro (60.9). This advantage is primarily attributed to our specialized Visual Checker module, which leverages the high-fidelity perception of Doubao-1.8 to rectify hallucinations and supplement granular visual details that generic models often overlook.

\begin{table}[t]
\centering
% \vspace{-0.4 cm}
\caption{Effect of different base models and DPO.}
\label{tab:dpo_and_basemodel}
\resizebox{\linewidth}{!}{%
\begin{tabular}{lcccc}
\toprule
\textbf{Model} & \textbf{Audio} & \textbf{Visual} & \textbf{AV Binding} & \textbf{AV Temporal} \\
\midrule
% AVoCaDO & 60.8    & 62.8    & 67.2    & 68.3    \\
Qwen2.5-Omni (SFT)     & 59.3    & 61.7    & 72.5    & 73.7    \\
Qwen2.5-Omni (SFT+DPO)     & 62.0    & 63.1    & 73.6    & 75.9    \\
Qwen3-Omni-30B (SFT)   & 61.2 & 63.6 & 73.2 & 76.9 \\
Qwen3-Omni-30B (SFT+DPO) & 63.0    & 66.8   & 75.3    & 79.2    \\
\bottomrule
\end{tabular}}
% \vspace{-0.7 cm}
    \vspace{-0.2cm}
\end{table}

\subsection{Exploring Another Base Models and DPO}
As shown in Table \ref{tab:dpo_and_basemodel}, we conduct additional experiments based on Qwen2.5-Omni, which similarly achieves competitive performance. Furthermore, we construct 2,000 DPO preference pairs based on our OCC pipeline. As demonstrated in Table \ref{tab:dpo_and_basemodel}, applying DPO to our SFT captioner significantly enhances performance across all dimensions. This validates the high quality of the OCC refinement data and proves its strong potential for automated preference alignment.

\subsection{Analysis for Our OCC Framework}
\begin{wrapfigure}{r}{0.5\textwidth} % r 代表靠右(Right)，0.5\textwidth 代表占据一半行宽
    \centering
    \includegraphics[width=1\linewidth]{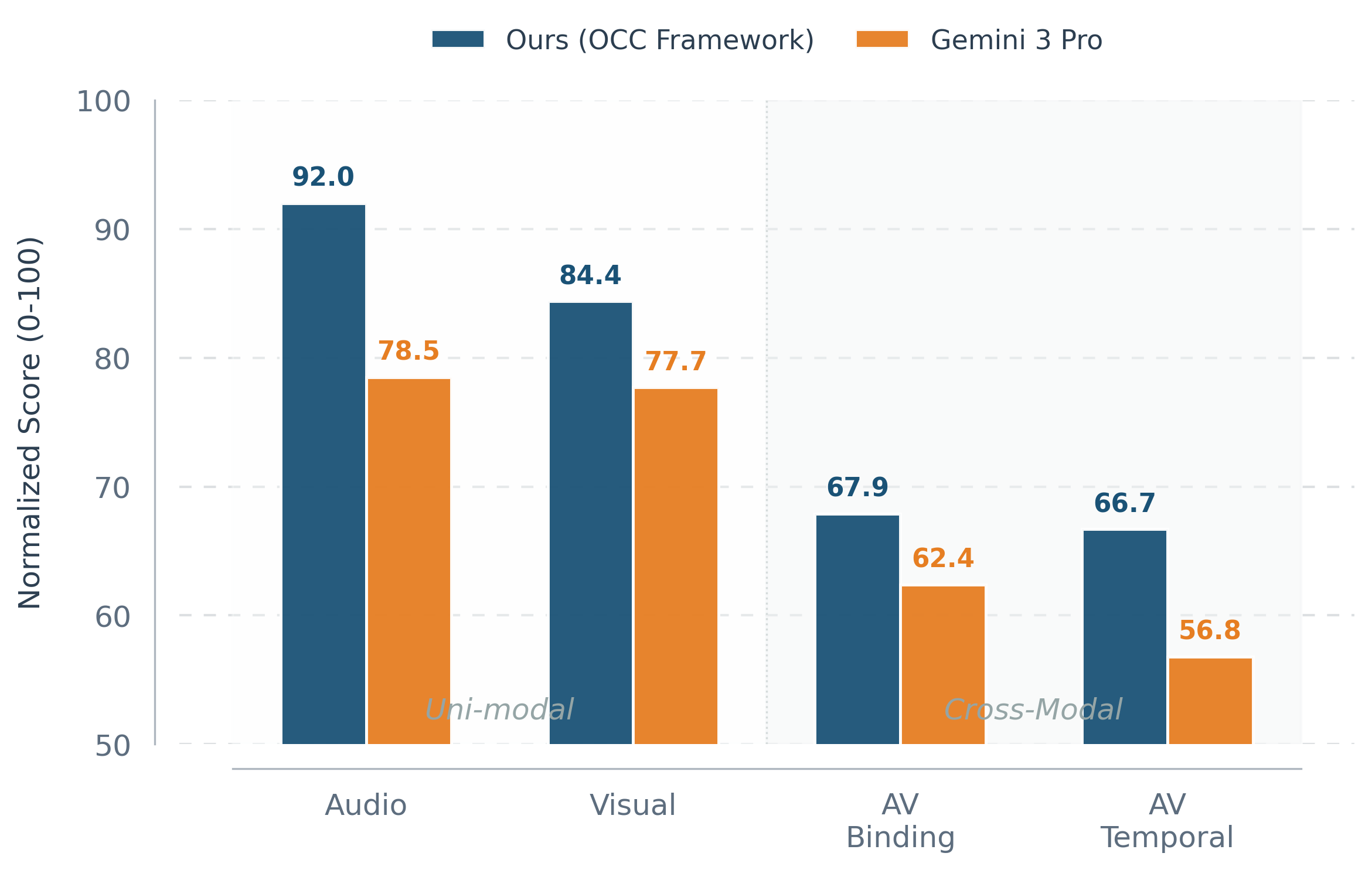}
    \vspace{-0.3cm} % 缩小图片下方间距
    \caption{The performance analysis for OCC framework.}
    \label{fig:abla}
    \vspace{-0.3cm} % 缩小标题下方间距
\end{wrapfigure}
Since the generation of LLM drafts in our benchmark construction utilized the OCC framework, a direct comparison within the full TCA-Bench would introduce inherent bias. To conduct a more rigorous and equitable ablation study of the OCC framework's contributions, we constructed an additional mini-bench comprising 20 samples. These samples utilized Gemini-2.5-Pro for initial draft generation and underwent stringent human auditing and manual refinement to ensure the same structural integrity as the main TCA-Bench.
As illustrated in Figure \ref{fig:abla}, we compared the OCC framework against  Gemini-3.0-Pro. The results demonstrate significant improvements across all evaluated dimensions.
The visual total score increased from 77.7 to 84.4, underscoring the critical role of the Visual Checker in refining visual details.
 The audio total score rose from 78.5 to 92.0, highlighting the substantial impact of the Audio Checker in rectifying auditory content.
While AV Temporal alignment saw a robust 9.9 improvement, the gain in AV Binding was more modest at 5.5.
This discrepancy suggests that for complex multi-person dialogue cases, severe hallucinations, particularly regarding the attribution of ambient sounds, persist even when the video is segmented into finer clips. Moreover, Table \ref{tab:Ablation} presents an ablation study where models are finetuned on data generated by specific OCC components. The results confirm that the Audio and Visual Checkers significantly boost their respective scores, while the A/V-Clip-Checker markedly enhances AV Binding and Temporal capabilities.

\begin{table}[t]
\centering
% \vspace{-0.2 cm}
\caption{Ablation study for the OCC framework.}
\label{tab:Ablation}
\resizebox{\linewidth}{!}{%
\begin{tabular}{lcccc}
\toprule
\textbf{Model} & \textbf{Audio} & \textbf{Visual} & \textbf{AV Binding} & \textbf{AV Temporal} \\
\midrule
% AVoCaDO & 60.8    & 62.8    & 67.2    & 68.3    \\
Observer  Only    &  57.6  & 56.1   &  66.5   &  69.8   \\
Observer + Audio  Checker    &  61.5   &  57.0   &  68.2   &   68.1    \\
Observer + Visual Checker    &  58.8   &  64.8   &  67.8   &    69.5    \\
Full OCC Framework (A/V-Clip-Checker) & 61.2    & 63.6   & 73.2    & 76.9   \\
\bottomrule
\end{tabular}}
% \vspace{-0.7 cm}
    \vspace{-0.2cm}
\end{table}
\section{Ethical Statement}
% We define our dataset and benchmark as a specialized collection of publicly available YouTube videos, meticulously paired with precise timestamps and fine-grained annotations. To ensure ethical compliance and minimize risk, the source material consists primarily of public-domain variety shows, films, and television series. The entire suite will be released under the CC BY-NC-SA 4.0 license, strictly limiting its application to non-commercial research and explicitly prohibiting uses related to facial recognition, mass surveillance, or biometric identification.
% To mitigate potential licensing conflicts and respect copyright boundaries, we do not distribute the raw video files directly. Instead, we provide the original YouTube URLs along with corresponding timestamps, supplemented by automated scripts for video downloading and clip extraction to facilitate community use. We remain committed to individual privacy; any person featured in the source videos may request the removal of their content, and we will promptly exclude the associated segments from all future releases.

Our dataset and benchmark comprise publicly available YouTube videos (e.g., variety shows and films) paired with precise timestamps and fine-grained annotations. %To ensure ethical compliance and respect copyright, we provide original URLs and extraction scripts.  %rather than raw video files. 
The suite will be released under a CC BY-NC-SA 4.0 license, strictly restricted to non-commercial research and explicitly prohibiting biometric or surveillance applications. Furthermore, we remain committed to privacy rights; any featured individual may request content removal, which we will fulfill by promptly excluding the relevant segments from future releases.

\section{Conclusion}
In this work, we present TCA-Captioner, a specialized framework designed to bridge the gap in  audiovisual video captioning. By integrating the iterative OCC framework and training on a curated high-density interaction dataset, our approach effectively alleviates modality detachment and temporal incoherence. Furthermore, the introduction of TCA-Bench enables a granular diagnostic evaluation of cross-modal interaction. Experimental results confirm that TCA-Captioner significantly advances the state-of-the-art in generating precise, temporally-aligned, and contextually rich audiovisual narratives. %More details are provided in the \textbf{Supplementary Material.}

\noindent \textbf{Acknowledgments.} This work was supported by Gusu Innovation and Entrepreneur Leading Talents: No. ZXL2024362, National Natural Science Foundation of China
: No. 62406135, Natural Science Foundation of Jiangsu Province: BK20241198, and Meituan.
% \clearpage\mbox{}Page \thepage\ of the manuscript.
% \clearpage\mbox{}Page \thepage\ of the manuscript.
% \clearpage\mbox{}Page \thepage\ of the manuscript.
% \clearpage\mbox{}Page \thepage\ of the manuscript.
% \clearpage\mbox{}Page \thepage\ of the manuscript. This is the last page.
% Now we have reached the maximum length of an ECCV \ECCVyear{} submission (excluding references and acknowledgements).
% References should start immediately after the main text, but can continue past p.\ 14 if needed. 
% \clearpage  % TODO FINAL: This \clearpage needs to be removed from both review and camera-ready versions.

% \section*{Acknowledgements}
% Please insert your acknowledgments here.

\bibliographystyle{splncs04}
\bibliography{main}

@String(CVPR  = {IEEE Conf. Comput. Vis. Pattern Recog.})

@String(AAAI  = {AAAI})

@String(ICASSP=	{ICASSP})

@String(CVPR  = {CVPR})

@article{zhang2026probing,
  title={Probing Effective and Efficient Category-Level Articulated Object Pose Perception},
  author={Zhang, Li and Meng, Xianhui and Liu, Liu and Jiang, Haonan and Wang, Jianan and Wang, Rujing and Lu, Cewu and Liu, Jun and Zhang, Hong},
  journal={IEEE Transactions on Pattern Analysis and Machine Intelligence},
  year={2026},
  publisher={IEEE}
}

@inproceedings{zhangrethinking,
  title={Rethinking 3D Convolution in $\ell_p$-norm Space},
  author={Zhang, Li and Zhong, Yan and Wang, Jianan and Min, Zhe and Liu, Liu and others},
  booktitle={The Thirty-eighth Annual Conference on Neural Information Processing Systems},
  year={2024}
}

@inproceedings{zhang2025gapt,
  title={GaPT-DAR: Category-level Garments Pose Tracking via Integrated 2D Deformation and 3D Reconstruction},
  author={Zhang, Li and Xu, Mingliang and Wang, Jianan and Yu, Qiaojun and Yang, Lixin and Li, Yonglu and Lu, Cewu and Wang, Rujing and Liu, Liu},
  booktitle={Proceedings of the Computer Vision and Pattern Recognition Conference},
  pages={22638--22647},
  year={2025}
}

@inproceedings{zhang2025u,
  title={U-COPE: Taking a Further Step to Universal 9D Category-Level Object Pose Estimation},
  author={Zhang, Li and Meng, Weiqing and Zhong, Yan and Kong, Bin and Xu, Mingliang and Du, Jianming and Wang, Xue and Wang, Rujing and Liu, Liu},
  booktitle={European Conference on Computer Vision},
  pages={254--270},
  year={2025},
  organization={Springer}
}

@article{wang2024qwen2,
  title={Qwen2-vl: Enhancing vision-language model's perception of the world at any resolution},
  author={Wang, Peng and Bai, Shuai and Tan, Sinan and Wang, Shijie and Fan, Zhihao and Bai, Jinze and Chen, Keqin and Liu, Xuejing and Wang, Jialin and Ge, Wenbin and others},
  journal={arXiv preprint arXiv:2409.12191},
  year={2024}
}

@article{zhu2023minigpt,
  title={Minigpt-4: Enhancing vision-language understanding with advanced large language models},
  author={Zhu, Deyao and Chen, Jun and Shen, Xiaoqian and Li, Xiang and Elhoseiny, Mohamed},
  journal={arXiv preprint arXiv:2304.10592},
  year={2023}
}

@inproceedings{zhang2023video,
  title={Video-llama: An instruction-tuned audio-visual language model for video understanding},
  author={Zhang, Hang and Li, Xin and Bing, Lidong},
  booktitle={Proceedings of the 2023 conference on empirical methods in natural language processing: system demonstrations},
  pages={543--553},
  year={2023}
}

@inproceedings{du2025vc4vg,
  title={VC4VG: Optimizing Video Captions for Text-to-Video Generation},
  author={Du, Yang and Lin, Zhuoran and Song, Kaiqiang and Wang, Biao and Zheng, Zhicheng and Ge, Tiezheng and Zheng, Bo and Jin, Qin},
  booktitle={Proceedings of the 2025 Conference on Empirical Methods in Natural Language Processing},
  pages={1124--1138},
  year={2025}
}

@article{chen2024sharegpt4video,
  title={Sharegpt4video: Improving video understanding and generation with better captions},
  author={Chen, Lin and Wei, Xilin and Li, Jinsong and Dong, Xiaoyi and Zhang, Pan and Zang, Yuhang and Chen, Zehui and Duan, Haodong and Tang, Zhenyu and Yuan, Li and others},
  journal={Advances in Neural Information Processing Systems},
  volume={37},
  pages={19472--19495},
  year={2024}
}

@inproceedings{fan2025instancecap,
  title={Instancecap: Improving text-to-video generation via instance-aware structured caption},
  author={Fan, Tiehan and Nan, Kepan and Xie, Rui and Zhou, Penghao and Yang, Zhenheng and Fu, Chaoyou and Li, Xiang and Yang, Jian and Tai, Ying},
  booktitle={Proceedings of the Computer Vision and Pattern Recognition Conference},
  pages={28974--28983},
  year={2025}
}

@inproceedings{lian2025describe,
  title={Describe anything: Detailed localized image and video captioning},
  author={Lian, Long and Ding, Yifan and Ge, Yunhao and Liu, Sifei and Mao, Hanzi and Li, Boyi and Pavone, Marco and Liu, Ming-Yu and Darrell, Trevor and Yala, Adam and others},
  booktitle={Proceedings of the IEEE/CVF International Conference on Computer Vision},
  pages={21766--21777},
  year={2025}
}

@article{qasim2025dense,
  title={Dense video captioning: A survey of techniques, datasets and evaluation protocols},
  author={Qasim, Iqra and Horsch, Alexander and Prasad, Dilip},
  journal={ACM Computing Surveys},
  volume={57},
  number={6},
  pages={1--36},
  year={2025},
  publisher={ACM New York, NY}
}

@article{guo2026alive,
  title={ALIVE: Animate Your World with Lifelike Audio-Video Generation},
  author={Guo, Ying and Gan, Qijun and Zhang, Yifu and Liu, Jinlai and Hu, Yifei and Xie, Pan and Qian, Dongjun and Zhang, Yu and Li, Ruiqi and Zhang, Yuqi and others},
  journal={arXiv preprint arXiv:2602.08682},
  year={2026}
}

@article{team2025longcat,
  title={Longcat-video technical report},
  author={Team, Meituan LongCat and Cai, Xunliang and Huang, Qilong and Kang, Zhuoliang and Li, Hongyu and Liang, Shijun and Ma, Liya and Ren, Siyu and Wei, Xiaoming and Xie, Rixu and others},
  journal={arXiv preprint arXiv:2510.22200},
  year={2025}
}

@inproceedings{wang2025haic,
  title={Haic: Improving human action understanding and generation with better captions for multi-modal large language models},
  author={Wang, Xiao and Hua, Jingyun and Lin, Weihong and Zhang, Yuanxing and Zhang, Fuzheng and Wu, Jianlong and Zhang, Di and Nie, Liqiang},
  booktitle={Proceedings of the 63rd Annual Meeting of the Association for Computational Linguistics (Volume 1: Long Papers)},
  pages={10158--10181},
  year={2025}
}

@inproceedings{kim2025vru,
  title={Vru-accident: A vision-language benchmark for video question answering and dense captioning for accident scene understanding},
  author={Kim, Younggun and Abdelrahman, Ahmed S and Abdel-Aty, Mohamed},
  booktitle={Proceedings of the IEEE/CVF International Conference on Computer Vision},
  pages={761--771},
  year={2025}
}

@article{yuan2025tarsier2,
  title={Tarsier2: Advancing large vision-language models from detailed video description to comprehensive video understanding},
  author={Yuan, Liping and Wang, Jiawei and Sun, Haomiao and Zhang, Yuchen and Lin, Yuan},
  journal={arXiv preprint arXiv:2501.07888},
  year={2025}
}

@article{xu2025qwen3,
  title={Qwen3-omni technical report},
  author={Xu, Jin and Guo, Zhifang and Hu, Hangrui and Chu, Yunfei and Wang, Xiong and He, Jinzheng and Wang, Yuxuan and Shi, Xian and He, Ting and Zhu, Xinfa and others},
  journal={arXiv preprint arXiv:2509.17765},
  year={2025}
}

@misc{gemini,
  author       = {{Gemini-3-Pro Team}},
  title        = {{Gemini-3-Pro}},
  howpublished = {\url{https://gemini.google.com}},
  note={Accessed: June 30, 2026}
}

@article{chen2025avocado,
  title={Avocado: An audiovisual video captioner driven by temporal orchestration},
  author={Chen, Xinlong and Ding, Yue and Lin, Weihong and Hua, Jingyun and Yao, Linli and Shi, Yang and Li, Bozhou and Zhang, Yuanxing and Liu, Qiang and Wan, Pengfei and others},
  journal={arXiv preprint arXiv:2510.10395},
  year={2025}
}

@inproceedings{nan2025openvid,
  title={Openvid-1m: A large-scale high-quality dataset for text-to-video generation},
  author={Nan, Kepan and Xie, Rui and Zhou, Penghao and Fan, Tiehan and Yang, Zhenheng and Chen, Zhijie and Li, Xiang and Yang, Jian and Tai, Ying},
  booktitle={International Conference on Learning Representations},
  volume={2025},
  pages={1045--1064},
  year={2025}
}

@misc{huang2026expo,
      title={Exposure Bias Can Alleviate Itself via Directional and Frequency Rectification in Flow Matching}, 
      author={Guanbo Huang and Jingjia Mao and Fanding Huang and Fengkai Liu and Xiangyang Luo and Yaoyuan Liang and Jiasheng Lu and Xiaoe Wang and Pei Liu and Ruiliu Fu and Ruqi Huang and Shao-Lun Huang},
       journal={arXiv preprint arXiv:2606.28226},
      year={2026},

}

@article{tang2025video,
  title={video-SALMONN 2: Caption-Enhanced Audio-Visual Large Language Models},
  author={Tang, Changli and Li, Yixuan and Yang, Yudong and Zhuang, Jimin and Sun, Guangzhi and Li, Wei and Ma, Zejun and Zhang, Chao},
  journal={arXiv preprint arXiv:2506.15220},
  year={2025}
}

@article{wu2025ugc,
  title={UGC-VideoCaptioner: An Omni UGC Video Detail Caption Model and New Benchmarks},
  author={Wu, Peiran and Liu, Yunze and Zhu, Zhengdong and Zhou, Enmin and Shen, Junxiao},
  journal={arXiv preprint arXiv:2507.11336},
  year={2025}
}

@inproceedings{liu2024improved,
  title={Improved baselines with visual instruction tuning},
  author={Liu, Haotian and Li, Chunyuan and Li, Yuheng and Lee, Yong Jae},
  booktitle={Proceedings of the IEEE/CVF conference on computer vision and pattern recognition},
  pages={26296--26306},
  year={2024}
}

@inproceedings{li2023blip,
  title={Blip-2: Bootstrapping language-image pre-training with frozen image encoders and large language models},
  author={Li, Junnan and Li, Dongxu and Savarese, Silvio and Hoi, Steven},
  booktitle={International conference on machine learning},
  pages={19730--19742},
  year={2023},
  organization={PMLR}
}

@inproceedings{chen2024internvl,
  title={Internvl: Scaling up vision foundation models and aligning for generic visual-linguistic tasks},
  author={Chen, Zhe and Wu, Jiannan and Wang, Wenhai and Su, Weijie and Chen, Guo and Xing, Sen and Zhong, Muyan and Zhang, Qinglong and Zhu, Xizhou and Lu, Lewei and others},
  booktitle={Proceedings of the IEEE/CVF conference on computer vision and pattern recognition},
  pages={24185--24198},
  year={2024}
}

@inproceedings{lin2024vila,
  title={Vila: On pre-training for visual language models},
  author={Lin, Ji and Yin, Hongxu and Ping, Wei and Molchanov, Pavlo and Shoeybi, Mohammad and Han, Song},
  booktitle={Proceedings of the IEEE/CVF conference on computer vision and pattern recognition},
  pages={26689--26699},
  year={2024}
}

@inproceedings{lin2024video,
  title={Video-llava: Learning united visual representation by alignment before projection},
  author={Lin, Bin and Ye, Yang and Zhu, Bin and Cui, Jiaxi and Ning, Munan and Jin, Peng and Yuan, Li},
  booktitle={Proceedings of the 2024 conference on empirical methods in natural language processing},
  pages={5971--5984},
  year={2024}
}

@article{zhang2024llava,
  title={Llava-video: Video instruction tuning with synthetic data},
  author={Zhang, Yuanhan and Wu, Jinming and Li, Wei and Li, Bo and Ma, Zejun and Liu, Ziwei and Li, Chunyuan},
  journal={arXiv preprint arXiv:2410.02713},
  year={2024}
}

@article{tang2023salmonn,
  title={Salmonn: Towards generic hearing abilities for large language models},
  author={Tang, Changli and Yu, Wenyi and Sun, Guangzhi and Chen, Xianzhao and Tan, Tian and Li, Wei and Lu, Lu and Ma, Zejun and Zhang, Chao},
  journal={arXiv preprint arXiv:2310.13289},
  year={2023}
}

@article{chu2024qwen2,
  title={Qwen2-audio technical report},
  author={Chu, Yunfei and Xu, Jin and Yang, Qian and Wei, Haojie and Wei, Xipin and Guo, Zhifang and Leng, Yichong and Lv, Yuanjun and He, Jinzheng and Lin, Junyang and others},
  journal={arXiv preprint arXiv:2407.10759},
  year={2024}
}

@article{hurst2024gpt,
  title={Gpt-4o system card},
  author={Hurst, Aaron and Lerer, Adam and Goucher, Adam P and Perelman, Adam and Ramesh, Aditya and Clark, Aidan and Ostrow, AJ and Welihinda, Akila and Hayes, Alan and Radford, Alec and others},
  journal={arXiv preprint arXiv:2410.21276},
  year={2024}
}

@misc{xu2025qwen25omnitechnicalreport,
      title={Qwen2.5-Omni Technical Report}, 
      author={Jin Xu and Zhifang Guo and Jinzheng He and Hangrui Hu and Ting He and Shuai Bai and Keqin Chen and Jialin Wang and Yang Fan and Kai Dang and Bin Zhang and Xiong Wang and Yunfei Chu and Junyang Lin},
      journal={arXiv preprint arXiv:2503.20215},
      year={2025},
      
}

@article{bai2025qwen3,
  title={Qwen3-vl technical report},
  author={Bai, Shuai and Cai, Yuxuan and Chen, Ruizhe and Chen, Keqin and Chen, Xionghui and Cheng, Zesen and Deng, Lianghao and Ding, Wei and Gao, Chang and Ge, Chunjiang and others},
  journal={arXiv preprint arXiv:2511.21631},
  year={2025}
}

@article{zhang2025videollama,
  title={Videollama 3: Frontier multimodal foundation models for image and video understanding},
  author={Zhang, Boqiang and Li, Kehan and Cheng, Zesen and Hu, Zhiqiang and Yuan, Yuqian and Chen, Guanzheng and Leng, Sicong and Jiang, Yuming and Zhang, Hang and Li, Xin and others},
  journal={arXiv preprint arXiv:2501.13106},
  year={2025}
}

@inproceedings{liu2024tempcompass,
  title={Tempcompass: Do video llms really understand videos?},
  author={Liu, Yuanxin and Li, Shicheng and Liu, Yi and Wang, Yuxiang and Ren, Shuhuai and Li, Lei and Chen, Sishuo and Sun, Xu and Hou, Lu},
  booktitle={Findings of the Association for Computational Linguistics: ACL 2024},
  pages={8731--8772},
  year={2024}
}

@article{zhong2025owlcap,
  title={Owlcap: Harmonizing motion-detail for video captioning via hmd-270k and caption set equivalence reward},
  author={Zhong, Chunlin and Hou, Qiuxia and Zhou, Zhangjun and Hao, Shuang and Lu, Haonan and Zhang, Yanhao and Tang, He and Bai, Xiang},
  journal={arXiv preprint arXiv:2508.18634},
  year={2025}
}

@article{wang2024tarsier,
  title={Tarsier: Recipes for training and evaluating large video description models},
  author={Wang, Jiawei and Yuan, Liping and Zhang, Yuchen and Sun, Haomiao},
  journal={arXiv preprint arXiv:2407.00634},
  year={2024}
}

@article{fu2024vita,
  title={Vita: Towards open-source interactive omni multimodal llm},
  author={Fu, Chaoyou and Lin, Haojia and Long, Zuwei and Shen, Yunhang and Dai, Yuhang and Zhao, Meng and Zhang, Yi-Fan and Dong, Shaoqi and Li, Yangze and Wang, Xiong and others},
  journal={arXiv preprint arXiv:2408.05211},
  year={2024}
}

@inproceedings{zhan2024anygpt,
  title={Anygpt: Unified multimodal llm with discrete sequence modeling},
  author={Zhan, Jun and Dai, Junqi and Ye, Jiasheng and Zhou, Yunhua and Zhang, Dong and Liu, Zhigeng and Zhang, Xin and Yuan, Ruibin and Zhang, Ge and Li, Linyang and others},
  booktitle={Proceedings of the 62nd Annual Meeting of the Association for Computational Linguistics (Volume 1: Long Papers)},
  pages={9637--9662},
  year={2024}
}

@inproceedings{lu2024unified,
  title={Unified-io 2: Scaling autoregressive multimodal models with vision language audio and action},
  author={Lu, Jiasen and Clark, Christopher and Lee, Sangho and Zhang, Zichen and Khosla, Savya and Marten, Ryan and Hoiem, Derek and Kembhavi, Aniruddha},
  booktitle={Proceedings of the IEEE/CVF Conference on Computer Vision and Pattern Recognition},
  pages={26439--26455},
  year={2024}
}

@article{nguyen2025see,
  title={See, Hear, and Understand: Benchmarking Audiovisual Human Speech Understanding in Multimodal Large Language Models},
  author={Nguyen, Le Thien Phuc and Yu, Zhuoran and Hang, Samuel Low Yu and An, Subin and Lee, Jeongik and Ban, Yohan and Chung, SeungEun and Nguyen, Thanh-Huy and Maeng, JuWan and Lee, Soochahn and others},
  journal={arXiv preprint arXiv:2512.02231},
  year={2025}
}

@inproceedings{su2023pandagpt,
  title={Pandagpt: One model to instruction-follow them all},
  author={Su, Yixuan and Lan, Tian and Li, Huayang and Xu, Jialu and Wang, Yan and Cai, Deng},
  booktitle={Proceedings of the 1st Workshop on Taming Large Language Models: Controllability in the era of Interactive Assistants!},
  pages={11--23},
  year={2023}
}

@article{zhang2025stream,
  title={Stream-omni: Simultaneous multimodal interactions with large language-vision-speech model},
  author={Zhang, Shaolei and Guo, Shoutao and Fang, Qingkai and Zhou, Yan and Feng, Yang},
  journal={arXiv preprint arXiv:2506.13642},
  year={2025}
}

@article{wang2025internvl3,
  title={Internvl3. 5: Advancing open-source multimodal models in versatility, reasoning, and efficiency},
  author={Wang, Weiyun and Gao, Zhangwei and Gu, Lixin and Pu, Hengjun and Cui, Long and Wei, Xingguang and Liu, Zhaoyang and Jing, Linglin and Ye, Shenglong and Shao, Jie and others},
  journal={arXiv preprint arXiv:2508.18265},
  year={2025}
}

@inproceedings{zhang2023speechgpt,
  title={Speechgpt: Empowering large language models with intrinsic cross-modal conversational abilities},
  author={Zhang, Dong and Li, Shimin and Zhang, Xin and Zhan, Jun and Wang, Pengyu and Zhou, Yaqian and Qiu, Xipeng},
  booktitle={Findings of the Association for Computational Linguistics: EMNLP 2023},
  pages={15757--15773},
  year={2023}
}

@article{abdar2024review,
  title={A review of deep learning for video captioning},
  author={Abdar, Moloud and Kollati, Meenakshi and Kuraparthi, Swaraja and Pourpanah, Farhad and McDuff, Daniel and Ghavamzadeh, Mohammad and Yan, Shuicheng and Mohamed, Abduallah and Khosravi, Abbas and Cambria, Erik and others},
  journal={IEEE Transactions on Pattern Analysis and Machine Intelligence},
  year={2024},
  publisher={IEEE}
}

@inproceedings{zhou2024streaming,
  title={Streaming dense video captioning},
  author={Zhou, Xingyi and Arnab, Anurag and Buch, Shyamal and Yan, Shen and Myers, Austin and Xiong, Xuehan and Nagrani, Arsha and Schmid, Cordelia},
  booktitle={Proceedings of the IEEE/CVF Conference on Computer Vision and Pattern Recognition},
  pages={18243--18252},
  year={2024}
}

@inproceedings{lin2022swinbert,
  title={Swinbert: End-to-end transformers with sparse attention for video captioning},
  author={Lin, Kevin and Li, Linjie and Lin, Chung-Ching and Ahmed, Faisal and Gan, Zhe and Liu, Zicheng and Lu, Yumao and Wang, Lijuan},
  booktitle={Proceedings of the IEEE/CVF conference on computer vision and pattern recognition},
  pages={17949--17958},
  year={2022}
}

@misc{tiktok_10m_2025,
  title={TikTok-10M: A Large-Scale Short Video Dataset for Video Understanding},
  author={The Data Company},
  year={2025},
  url={https://huggingface.co/datasets/The-data-company/TikTok-10M},
note={Accessed: June 30, 2026}
}

@misc{SileroVAD,
  author = {Silero Team},
  title = {Silero VAD: pre-trained enterprise-grade Voice Activity Detector (VAD), Number Detector and Language Classifier},
  year = {2024},
  publisher = {GitHub},
  journal = {GitHub repository}
}

@inproceedings{varghese2024yolov8,
  title={Yolov8: A novel object detection algorithm with enhanced performance and robustness},
  author={Varghese, Rejin and Sambath, M},
  booktitle={2024 International conference on advances in data engineering and intelligent computing systems (ADICS)},
  pages={1--6},
  year={2024},
  organization={IEEE}
}

@misc{Farré2024FineVideo,
  title={FineVideo},
  author={Farré, Miquel and Marafioti, Andi and Tunstall, Lewis and Von Werra, Leandro and Wolf, Thomas},
  year={2024},
  howpublished={\url{https://huggingface.co/datasets/HuggingFaceFV/finevideo}},
  note={Accessed: June 30, 2026}
}

@inproceedings{shang2025large,
  title={A large-scale dataset with behavior, attributes, and content of mobile short-video platform},
  author={Shang, Yu and Gao, Chen and Li, Nian and Li, Yong},
  booktitle={Companion Proceedings of the ACM on Web Conference 2025},
  pages={793--796},
  year={2025}
}

@article{ma2025omni,
  title={Omni-Captioner: Data Pipeline, Models, and Benchmark for Omni Detailed Perception},
  author={Ma, Ziyang and Xu, Ruiyang and Xing, Zhenghao and Chu, Yunfei and Wang, Yuxuan and He, Jinzheng and Xu, Jin and Heng, Pheng-Ann and Yu, Kai and Lin, Junyang and others},
  journal={arXiv preprint arXiv:2510.12720},
  year={2025}
}

@article{yang2025humanomniv2,
  title={Humanomniv2: From understanding to omni-modal reasoning with context},
  author={Yang, Qize and Yao, Shimin and Chen, Weixuan and Fu, Shenghao and Bai, Detao and Zhao, Jiaxing and Sun, Boyuan and Yin, Bowen and Wei, Xihan and Zhou, Jingren},
  journal={arXiv preprint arXiv:2506.21277},
  year={2025}
}

@article{ge2025arc,
  title={Arc-hunyuan-video-7b: Structured video comprehension of real-world shorts},
  author={Ge, Yuying and Ge, Yixiao and Li, Chen and Wang, Teng and Pu, Junfu and Li, Yizhuo and Qiu, Lu and Ma, Jin and Duan, Lisheng and Zuo, Xinyu and others},
  journal={arXiv preprint arXiv:2507.20939},
  year={2025}
}

@article{chen2023diffusion,
  title={Diffusion model for camouflaged object detection},
  author={Chen, Zhennan and Gao, Rongrong and Xiang, Tian-Zhu and Lin, Fan},
  journal={arXiv preprint arXiv:2308.00303},
  year={2023}
}

@inproceedings{chen2025ragd,
  title={RAGD: Regional-Aware Diffusion Model for Text-to-Image Generation},
  author={Chen, Zhennan and Li, Yajie and Wang, Haofan and Chen, Zhibo and Jiang, Zhengkai and Li, Jun and Wang, Qian and Yang, Jian and Tai, Ying},
  booktitle={Proceedings of the IEEE/CVF International Conference on Computer Vision},
  pages={19331--19341},
  year={2025}
}

@article{zhou20243dis,
  title={3dis: Depth-driven decoupled instance synthesis for text-to-image generation},
  author={Zhou, Dewei and Xie, Ji and Yang, Zongxin and Yang, Yi},
  journal={arXiv preprint arXiv:2410.12669},
  year={2024}
}

@article{zhou2025dreamrenderer,
  title={Dreamrenderer: Taming multi-instance attribute control in large-scale text-to-image models},
  author={Zhou, Dewei and Li, Mingwei and Yang, Zongxin and Yang, Yi},
  journal={arXiv preprint arXiv:2503.12885},
  year={2025}
}

@article{zhou2025bidedpo,
  title={BideDPO: Conditional Image Generation with Simultaneous Text and Condition Alignment},
  author={Zhou, Dewei and Li, Mingwei and Yang, Zongxin and Lu, Yu and Xu, Yunqiu and Wang, Zhizhong and Huang, Zeyi and Yang, Yi},
  journal={arXiv preprint arXiv:2511.19268},
  year={2025}
}

@article{zhou2026metapoint,
  title={MetaPoint: Unlocking Precise Spatial Control in Agentic Visual Generation},
  author={Zhou, Dewei and Huang, Xinyu and Wang, Xun and Xie, Ji and Zhang, Yabo and Li, Liang and Li, Kunchang and Yang, Zongxin and Yang, Yi},
  journal={arXiv preprint arXiv:2606.05031},
  year={2026}
}

@inproceedings{lu2024mace,
  title={Mace: Mass concept erasure in diffusion models},
  author={Lu, Shilin and Wang, Zilan and Li, Leyang and Liu, Yanzhu and Kong, Adams Wai-Kin},
  booktitle={Proceedings of the IEEE/CVF Conference on Computer Vision and Pattern Recognition},
  pages={6430--6440},
  year={2024}
}

@inproceedings{lu2023tf,
  title={Tf-icon: Diffusion-based training-free cross-domain image composition},
  author={Lu, Shilin and Liu, Yanzhu and Kong, Adams Wai-Kin},
  booktitle={Proceedings of the IEEE/CVF International Conference on Computer Vision},
  pages={2294--2305},
  year={2023}
}

@article{lu2024robust,
  title={Robust watermarking using generative priors against image editing: From benchmarking to advances},
  author={Lu, Shilin and Zhou, Zihan and Lu, Jiayou and Zhu, Yuanzhi and Kong, Adams Wai-Kin},
  journal={arXiv preprint arXiv:2410.18775},
  year={2024}
}

@inproceedings{zhao2024toward,
  title={Toward sufficient spatial-frequency interaction for gradient-aware underwater image enhancement},
  author={Zhao, Chen and Cai, Weiling and Dong, Chenyu and Zeng, Ziqi},
  booktitle={ICASSP 2024-2024 IEEE International Conference on Acoustics, Speech and Signal Processing (ICASSP)},
  pages={3220--3224},
  year={2024},
  organization={IEEE}
}

@article{zhao2026spiking,
  title={Spiking pyramid wavelet transformation for high-efficient and low-energy image restoration},
  author={Zhao, Chen and Hu, Xiantao and Wu, Song and Wang, Qian and Wu, Chen and Xie, Rui and Yang, Jian and Tai, Ying},
  journal={Pattern Recognition},
  pages={114297},
  year={2026},
  publisher={Elsevier}
}

@article{zhou2026learning,
  title={Learning to Refine: Spectral-Decoupled Iterative Refinement Framework for Precipitation Nowcasting},
  author={Zhou, Yunlong and Zhao, Chen and Peng, Danyang and Ji, Fanfan and Yuan, Xiao-Tong},
  journal={arXiv preprint arXiv:2606.02661},
  year={2026}
}

@inproceedings{zhao2025zero,
  title={From zero to detail: Deconstructing ultra-high-definition image restoration from progressive spectral perspective},
  author={Zhao, Chen and Chen, Zhizhou and Xu, Yunzhe and Gu, Enxuan and Li, Jian and Yi, Zili and Wang, Qian and Yang, Jian and Tai, Ying},
  booktitle={Proceedings of the IEEE/CVF Conference on Computer Vision and Pattern Recognition},
  pages={17935--17946},
  year={2025}
}

@inproceedings{hu2025exploiting,
  title={Exploiting multimodal spatial-temporal patterns for video object tracking},
  author={Hu, Xiantao and Tai, Ying and Zhao, Xu and Zhao, Chen and Zhang, Zhenyu and Li, Jun and Zhong, Bineng and Yang, Jian},
  booktitle={Proceedings of the AAAI Conference on Artificial Intelligence},
  volume={39},
  number={4},
  pages={3581--3589},
  year={2025}
}

@article{xie2024addsr,
  title={Addsr: Accelerating diffusion-based blind super-resolution with adversarial diffusion distillation},
  author={Xie, Rui and Zhao, Chen and Zhang, Kai and Zhang, Zhenyu and Zhou, Jun and Yang, Jian and Tai, Ying},
  journal={arXiv preprint arXiv:2404.01717},
  year={2024}
}

@article{zhao2025spectral,
  title={Spectral normalization and dual contrastive regularization for image-to-image translation},
  author={Zhao, Chen and Cai, Wei-Ling and Yuan, Zheng},
  journal={The Visual Computer},
  volume={41},
  number={1},
  pages={129--140},
  year={2025},
  publisher={Springer}
}

@article{zhao2024cycle,
  title={Cycle contrastive adversarial learning with structural consistency for unsupervised high-quality image deraining transformer},
  author={Zhao, Chen and Cai, Weiling and Hu, Chengwei and Yuan, Zheng},
  journal={Neural Networks},
  volume={178},
  pages={106428},
  year={2024},
  publisher={Elsevier}
}

@inproceedings{dong2025mamba,
  title={O-mamba: O-shape state-space model for underwater image enhancement},
  author={Dong, Chenyu and Zhao, Chen and Cai, Weiling and Yang, Bo and Guo, Yuqing},
  booktitle={Chinese Conference on Pattern Recognition and Computer Vision (PRCV)},
  pages={168--182},
  year={2025},
  organization={Springer}
}

@article{zhao2025multi,
  title={Multi-cropping contrastive learning and domain consistency for unsupervised image-to-image translation},
  author={Zhao, Chen and Cai, Wei-Ling and Yuan, Zheng and Hu, Cheng-Wei},
  journal={IET Image Processing},
  volume={19},
  number={1},
  pages={e70006},
  year={2025},
  publisher={Wiley Online Library}
}

@article{zhao2025learning,
  title={Learning Multi-scale Spatial-frequency Features for Image Denoising},
  author={Zhao, Xu and Zhao, Chen and Hu, Xiantao and Zhang, Hongliang and Tai, Ying and Yang, Jian},
  journal={arXiv preprint arXiv:2506.16307},
  year={2025}
}

@article{zhou2026more,
  title={More realistic and accurate precipitation nowcasting with Conditional Rectified Flow Transformers},
  author={Zhou, Yunlong and Zhao, Chen and Ji, Fanfan and Hang, Renlong and Liu, Qingshan and Yuan, Xiao-Tong},
  journal={Engineering Applications of Artificial Intelligence},
  volume={165},
  pages={113402},
  year={2026},
  publisher={Elsevier}
}

@article{lu2025does,
  title={Does flux already know how to perform physically plausible image composition?},
  author={Lu, Shilin and Lian, Zhuming and Zhou, Zihan and Zhang, Shaocong and Zhao, Chen and Kong, Adams Wai-Kin},
  journal={arXiv preprint arXiv:2509.21278},
  year={2025}
}

@ARTICLE{11513016,
  author={Zhao, Chen and Xu, Yunzhe and Chen, Zhizhou and Gu, Enxuan and Zhang, Kai and Liu, Xiaoming and Yang, Jian and Tai, Ying},
  journal={IEEE Transactions on Pattern Analysis and Machine Intelligence}, 
  title={From Zero to Detail: A Progressive Spectral Decoupling Paradigm for UHD Image Restoration with New Benchmark}, 
  year={2026},
  pages={1-18},
}

@inproceedings{Zhaowfdiff,
  author={Zhao, Chen and Cai, Weiling and Dong, Chenyu and Hu, Chengwei},
  title={Wavelet-based Fourier Information Interaction with Frequency Diffusion Adjustment for Underwater Image Restoration},
  booktitle={Proceedings of the IEEE/CVF Conference on Computer Vision and Pattern Recognition (CVPR)},
  month={June},
  year={2024}
}

@article{zhao2026learning,
  title={Learning a physical-aware diffusion model based on transformer for underwater image enhancement},
  author={Zhao, Chen and Dong, Chenyu and Cai, Weiling and Wang, Yueyue},
  journal={IEEE Transactions on Geoscience and Remote Sensing},
  year={2026},
  publisher={IEEE}
}

@article{zhao2025ultrahr,
  title={UltraHR-100K: Enhancing UHR Image Synthesis with A Large-Scale High-Quality Dataset},
  author={Zhao, Chen and Ci, En and Xu, Yunzhe and Fan, Tiehan and Guan, Shanyan and Ge, Yanhao and Yang, Jian and Tai, Ying},
  journal={Advances in Neural Information Processing Systems},
  year={2025}
}

@article{liu2026driveva,
  title={Driveva: Video action models are zero-shot drivers},
  author={Liu, Mengmeng and Zhang, Diankun and Liu, Jiuming and Cui, Jianfeng and Xie, Hongwei and Chen, Guang and Ye, Hangjun and Yang, Michael Ying and Nex, Francesco and Cheng, Hao},
  journal={arXiv preprint arXiv:2604.04198},
  year={2026}
}

@article{chen2025dip,
  title={Dip: Taming diffusion models in pixel space},
  author={Chen, Zhennan and Zhu, Junwei and Chen, Xu and Zhang, Jiangning and Hu, Xiaobin and Zhao, Hanzhen and Wang, Chengjie and Yang, Jian and Tai, Ying},
  journal={arXiv preprint arXiv:2511.18822},
  year={2025}
}

@article{chen2026l2punlockinglatentpotential,
  title={L2P: Unlocking Latent Potential for Pixel Generation},
  author={Chen, Zhennan and Zhu, Junwei and Chen, Xu and Zhang, Jiangning and Chen, Jiawei and Zeng, Zhuoqi and Zhang, Wei and Wang, Chengjie and Yang, Jian and Tai, Ying},
  journal={arXiv preprint arXiv:2605.12013},
  year={2026}
}

@article{chen2026echoefficientchestxray,
  title={ECHO: Efficient Chest X-ray Report Generation with One-step Block Diffusion},
  author={Chen, Lifeng and You, Tianqi and Liu, Hao and Bao, Zhimin and Jiao, Jile and Han, Xiao and Ou, Zhicai and Sun, Tao and Mou, Xiaofeng and Jin, Xiaojie and others},
  journal={arXiv preprint arXiv:2604.09450},
  year={2026}
}

@ARTICLE{11551836,
  author={Chen, Lifeng and Wang, Jiner and Pan, Zihao and Zhu, Beier and Yang, Xiaofeng and Zhang, Chi},
  journal={IEEE Transactions on Image Processing}, 
  title={Detail++: Training-Free Detail Enhancer for T2I Diffusion Models}, 
  year={2026},
  volume={35},
  number={},
  pages={5982-5994}}

@article{du2026unsupervised,
  title={Unsupervised Hyperspectral Image Super-Resolution via Self-Supervised Modality Decoupling},
  author={Du, Songcheng and Zou, Yang and Wang, Zixu and Li, Xingyuan and Li, Ying and Shang, Changjing and Shen, Qiang},
  journal={International Journal of Computer Vision},
  volume={134},
  pages={152},
  year={2026},
  publisher={Springer Nature}
}

@inproceedings{du2026pansharpening,
  title={Pansharpening for thin-cloud contaminated remote sensing images: a unified framework and benchmark dataset},
  author={Du, Songcheng and Zou, Yang and Li, Jiaxin and Liu, Mingxuan and Li, Ying and Shang, Changjing and Shen, Qiang},
  booktitle={Proceedings of the AAAI Conference on Artificial Intelligence},
  volume={40},
  number={5},
  pages={3696--3704},
  year={2026}
}

@inproceedings{li2026unifusion,
  title={Unifusion: A unified image fusion framework with robust representation and source-aware preservation},
  author={Li, Xingyuan and Du, Songcheng and Zou, Yang and Xu, HaoYuan and Jiang, Zhiying and Liu, Jinyuan},
  booktitle={Proceedings of the IEEE/CVF Conference on Computer Vision and Pattern Recognition},
  pages={33869--33880},
  year={2026}
}

@article{jiang2026imagine,
  title={Imagine Before You Predict: Interleaved Latent Visual Reasoning for Video Event Prediction},
  author={Jiang, Tianxiang and Wu, Linquan and Xia, Sheng and Li, Songze and Yan, Ziang and Yang, Haoyu and Qiao, Yu and Wang, Yi},
  journal={arXiv preprint arXiv:2606.05769},
  year={2026}
}

@article{jiang2025vknowu,
  title={VKnowU: Evaluating Visual Knowledge Understanding in Multimodal LLMs},
  author={Jiang, Tianxiang and Xia, Sheng and Xu, Yicheng and Wu, Linquan and Zeng, Xiangyu and Wang, Limin and Qiao, Yu and Wang, Yi},
  journal={arXiv preprint arXiv:2511.20272},
  year={2025}
}

@article{chen2026calibrated,
  title={Calibrated Harmonic Overlaid Implicit Neural Representations for Multi-Dimensional Data},
  author={Chen, Honghang and Zhang, Xiujun and Sun, Xiaoli and Xiao Mingqing},
  journal={arXiv preprint arXiv:2606.26763},
  year={2026}
}

@article{yao2024minicpm,
  title={MiniCPM-V: A GPT-4V Level MLLM on Your Phone},
  author={Yao, Yuan and Yu, Tianyu and Zhang, Ao and Wang, Chongyi and Cui, Junbo and Zhu, Hongji and Cai, Tianchi and Li, Haoyu and Zhao, Weilin and He, Zhihui and others},
  journal={arXiv preprint arXiv:2408.01800},
  year={2024}
}

@article{zhao2026luve,
  title={LUVE: Latent-Cascaded Ultra-High-Resolution Video Generation with Dual Frequency Experts},
  author={Zhao, Chen and Chen, Jiawei and Li, Hongyu and Kang, Zhuoliang and Lu, Shilin and Wei, Xiaoming and Zhang, Kai and Yang, Jian and Tai, Ying},
  journal={arXiv preprint arXiv:2602.11564},
  year={2026}
}
\end{document}